\documentclass[10pt,twocolumn,letterpaper]{article}

\usepackage{wacv}              % To produce the CAMERA-READY version
\usepackage{graphicx}
\usepackage{threeparttable}
\usepackage{float}

\usepackage{amsmath}
\usepackage{amssymb}
\usepackage[utf8]{inputenc}
\usepackage{multirow}
\usepackage{booktabs}
\usepackage{arydshln}
\usepackage{enumitem}
\makeatletter
\def\adl@drawivfalse{} % to prevent compile error if needed
\makeatother

\usepackage[pagebackref,breaklinks,colorlinks]{hyperref}

\usepackage[capitalize]{cleveref}
\crefname{section}{Sec.}{Secs.}
\Crefname{section}{Section}{Sections}
\Crefname{table}{Table}{Tables}
\crefname{table}{Tab.}{Tabs.}

\def\wacvPaperID{1008} % *** Enter the WACV Paper ID here
\def\confName{WACV}
\def\confYear{2026}

\begin{document}

\title{Non-Aligned Reference Image Quality Assessment for Novel View Synthesis}

\author{
Abhijay Ghildyal$^{1}$, Rajesh Sureddi$^{2}$, Nabajeet Barman$^{1}$, Saman Zadtootaghaj$^{1}$, Alan Bovik$^{2}$ \\
$^{1}$Sony Interactive Entertainment \quad $^{2}$University of Texas at Austin \\
}
% email is not necessary
% WACV camera-ready instructions explicitly state:
% "The ABSTRACT is to be in fully justified italicized text, at the top of the left-hand column, below the author and affiliation information."
% This implies the author information under the title should include only the name and affiliation, not email. 

% \and
\maketitle

%%%%%%%%% ABSTRACT
\begin{abstract}
Evaluating the perceptual quality of Novel View Synthesis (NVS) images remains a key challenge, particularly in the absence of pixel-aligned ground truth references. Full-Reference Image Quality Assessment (FR-IQA) methods fail under misalignment, while No-Reference (NR-IQA) methods struggle with generalization. In this work, we introduce a Non-Aligned Reference (NAR-IQA) framework tailored for NVS, where it is assumed that the reference view shares partial scene content but lacks pixel-level alignment. We constructed a large-scale image dataset containing synthetic distortions targeting Temporal Regions of Interest (TROI) to train our NAR-IQA model. Our model is built on a contrastive learning framework that incorporates LoRA-enhanced DINOv2 embeddings and is guided by supervision from existing IQA methods. We train exclusively on synthetically generated distortions, deliberately avoiding overfitting to specific real NVS samples and thereby enhancing the model’s generalization capability. Our model outperforms state-of-the-art FR-IQA, NR-IQA, and NAR-IQA methods, achieving robust performance on both aligned and non-aligned references. We also conducted a novel user study to gather data on human preferences when viewing non-aligned references in NVS. We find strong correlation between our proposed quality prediction model and the collected subjective ratings. For dataset, and code, please visit our project page: \url{https://stootaghaj.github.io/nova-project/}

% human judgments.
% across 17 diverse NVS scenes

\end{abstract}

\section{Introduction}
\label{sec:intro}

\begin{figure*}[t]
    \centering
    \includegraphics[width=0.98\linewidth]{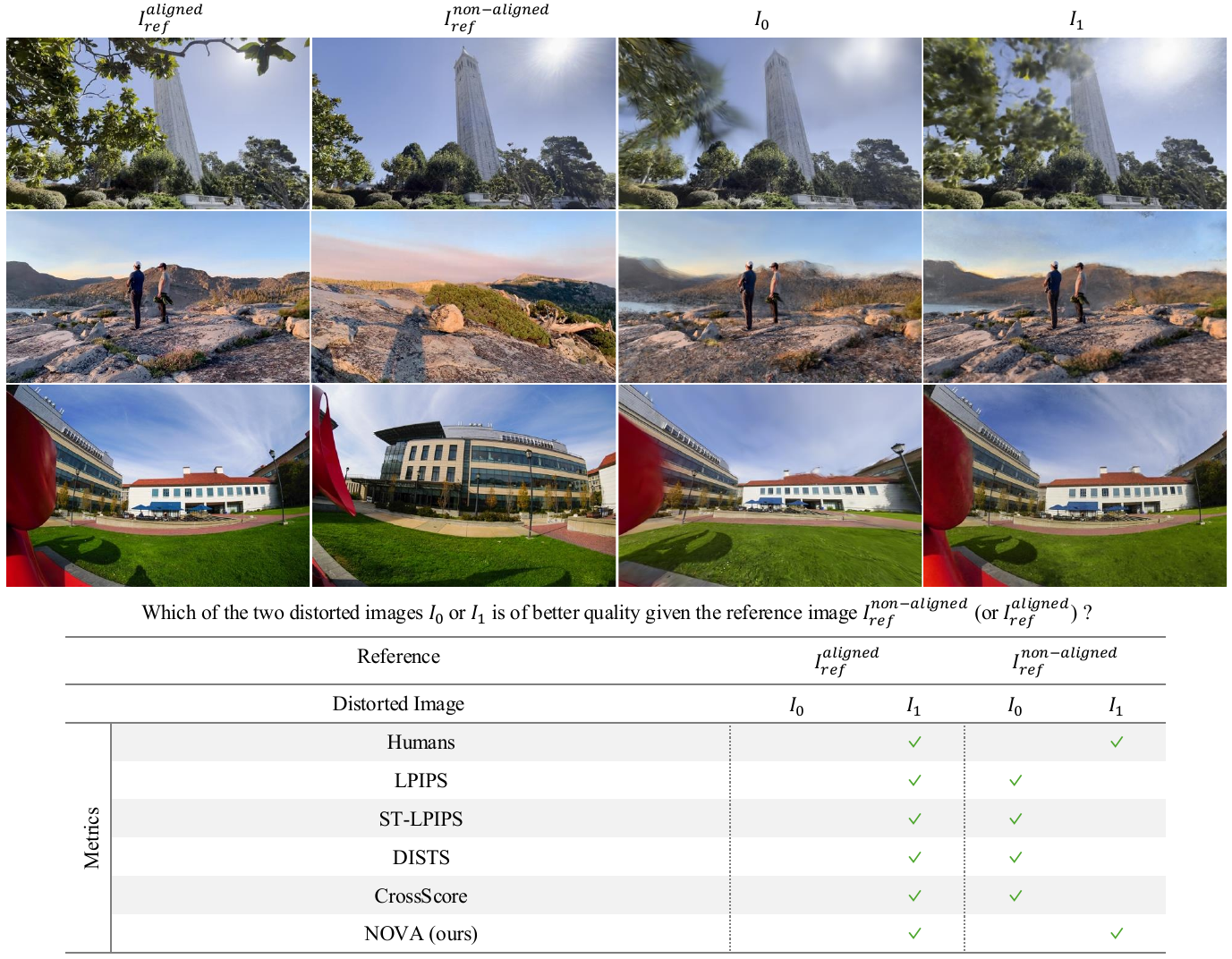}
    \vspace{-0.1in}
    \caption{
    Visual comparison of aligned and non-aligned references for assessing the quality of distorted images \( I_0 \) and \( I_1 \). Each row shows a sample from the dataset with an aligned reference, a non-aligned reference, and two distorted views. The table below highlights metric performance (e.g., LPIPS, DISTS, CrossScore) and human preferences under both reference settings.
    }
    \label{fig:teaser}
    \vspace{-0.15in}
\end{figure*}

% Image Quality Assessment (IQA) plays a crucial role in the development and evaluation of image generation models by providing quantitative measures of visual fidelity. 
Image Quality Assessment (IQA) plays a crucial role in enabling perceptually accurate benchmarking, comparison, and optimization of models across a wide range of vision tasks. IQA methods can be broadly categorized into four types. Full-reference (FR) methods require high-quality reference images to assess image degradations by pixelwise comparisons or in a perceptual feature space~\cite{wang2003multiscale,wang2004image,zhang2018perceptual,Ding20,andersson2020flip,ghildyal2022stlpips,ghildyal2025foundation}. Reduced-reference (RR) methods, designed for scenarios where the full reference is unavailable or impractical to transmit, use partial information extracted from the reference image, offering trade-offs between accuracy and applicability~\cite{wang2005reduced,li2009reduced,soundararajan2011rred,wang2016reduced,yang2022maniqa}. No-reference (NR) methods, on the other hand, evaluate image quality without access to any reference, making them suitable for scenarios where ground truth is unavailable~\cite{agnolucci2024arniqa,madhusudana2022image,jamshidi2025lar,Su_2020_CVPR,sureddi2025triqa,wang2024blind,sun2022graphiqa}. Non-aligned reference (NAR) IQA methods bridge the gap between FR and NR approaches by enabling quality assessment using a reference that shares content with the test image, but without requiring precise spatial alignment~\cite{liang2016image,yin2022content,li2023less,shi2023robust}. Unlike FR-IQA methods, NAR-IQA methods do not rely on strict pixel-to-pixel or region-level correspondences.

Recent efforts to evaluate synthesized novel views from Neural Radiance Fields (NeRFs) have highlighted that the visual artifacts and misalignments generated by these models differ markedly from distortions typically found in conventional IQA datasets~\cite{zhou2023nerflix,pedronvs,zhang2025evaluating}. As a result, existing learned IQA models, which are often trained on traditional degradations, struggle to generalize effectively to this context~\cite{zhang2025evaluating,pedronvs}. This challenge is not unique to NeRFs; even in standard IQA and video quality assessment (VQA) pipelines, perfect reference images are not always accessible. In such cases, practitioners frequently resort to no-reference (NR-IQA) models that assess quality based only on distorted inputs. However, real-world applications often offer a compromise: non-aligned but semantically similar reference images, such as a representative frame or alternate view of the same scene. Our approach leverages this setting through Non-Aligned Reference IQA (NAR-IQA), which incorporates contextual cues from imperfect references without requiring spatial alignment. \textit{In practical scenarios where perfectly aligned reference images are rare or unavailable, NAR-IQA offers a more realistic and robust pathway for perceptual quality assessment.}

Among existing efforts toward evaluating novel views, commonly referred to as Novel View Synthesis Quality Assessment (NVS QA)~\cite{pedronvs,martin2025gs,explicitnqa,tabassum2024quality,zhang2025evaluating,qunerfnqa,qu2025nvs}, CrossScore introduces a strategy that leverages non-aligned reference views~\cite{wang2024crossscore}. It introduces the concept of cross-reference IQA, a paradigm wherein multiple non-aligned reference views are leveraged to assess image quality more effectively. In their approach, the SSIM value computed between an original reference image and a distorted version of it is treated as ground truth. A model is trained to replicate this score using several non-aligned views as reference. While this approach works under certain conditions, it has two key drawbacks: it relies on NVS-generated data, which can bias the model toward specific distortions, and it uses SSIM as a target. Building on this foundation, our work takes a different direction by not using SSIM as a training signal, since studies show it has limited correlation with human perceptual judgments as compared to more recent learned models~\cite{ghildyal2022stlpips,zhang2018perceptual,ghildyal2025foundation,Ding20,sangnie2020,ghildyal2023attacking,prashnani2018pieapp}. We selected DeepDC~\cite{zhu2023DeepDC} and DISTS \cite{Ding20} scores as more appropriate training targets, given their higher alignment with subjective quality assessments. To ensure that our learned model generalizes beyond the specific artifacts introduced by NVS methods, we incorporate contrastive learning on a diverse collection of localized synthetic distortions, encouraging the model to learn discriminative features without overfitting to only certain NVS distortion types. Our main contributions are as follows:
\begin{enumerate}[topsep=0pt, partopsep=0pt, itemsep=3pt, parsep=0pt]
    \item We present the first subjective study of NAR-IQA in the context of NVS-QA.
    \item We construct a diverse dataset by training NeRF/GS models with varied settings, capturing a range of distortion severities and types across 17 scenes, along with varying levels of spatial overlap between reference and distorted images.
    \item We propose a supervised contrastive learning-based framework that leverages synthetic distortions and supervision from existing FR-IQA metrics to train a robust NAR-IQA model.
    \item We evaluate our model, which we call Non-aligned View Assessment (NOVA), on our new benchmark and on the existing NVS-QA dataset~\cite{pedronvs} to assess its generalization capability.
\end{enumerate}

\section{Related Work}
\label{sec:related}

\subsection{Non-aligned Reference IQA}

% appears abrupt - I would say something like FR-IQA metrics such as SSIM and MS-SSIM uses two Aligned references. For example, SSIM compares two images...
The Structural Similarity (SSIM) Index is a widely adopted perceptual method of image quality assessment (IQA), designed to compare two images by evaluating structural, luminance, and contrast similarities. Due to its reliance on local patch-wise computations via a sliding window, SSIM assumes spatial alignment between pairs of compared images. A multi-scale extension, MS-SSIM, addresses this limitation to some extent by evaluating image similarity over multiple spatial resolutions, making it more robust to small misalignments. However, both SSIM and MS-SSIM assume pixel-level correspondence, an assumption often violated in NVS, where synthesized views may be geometrically misaligned with references or differ in content due to occlusions, viewpoint changes, or synthesis artifacts—making strict spatial comparisons unreliable.

% However, both SSIM and MS-SSIM are fundamentally predicated on the assumption of pixel-level correspondence, which is often violated in novel view synthesis (NVS) scenarios, where the generated view rarely aligns closely with any available reference images. In these cases, the generated view may not only be geometrically misaligned with available references but may also differ in visible content due to occlusions, viewpoint shifts, or synthesis artifacts—rendering strict spatial comparison unreliable

% However, both SSIM and MS-SSIM are fundamentally predicated on the assumption of pixel-level correspondence, which is often violated in novel view synthesis (NVS) scenarios, where the generated view rarely aligns closely with any available reference images.

% To address such misalignments, recent research has focused on non-aligned references (NAR-IQA), primarily outside the NVS domain, a paradigm in which the reference image does not share precise spatial or semantic alignment with the distorted image. 

To address such misalignments, although primarily outside the NVS domain, recent research has focused on NAR-IQA, a paradigm wherein the reference images lack precise spatial or semantic alignment with the distorted images. Yin et al.~\cite{yin2022content} introduced the Content-Variant Reference Knowledge Distillation for IQA (CVRKD) framework, where the high-quality (HQ) reference images do not depict the same scene content as the low-quality (LQ) image. In this approach, an FR-IQA model serves as a teacher network, while a student network is trained in an offline manner using randomly sampled patches from unrelated HQ images. Despite being trained under NAR conditions, the student model achieves performance comparable to the teacher when evaluated on aligned image pairs, demonstrating the robustness imparted by knowledge distillation. Similarly, Li et al.~\cite{li2023less} proposed a framework that extends the NAR-IQA concept towards the NR regime. In their model, the teacher network learns from HQ images lacking spatial or semantic alignment with the LQ images, while the student is trained exclusively in an NR setting. This hybrid design benefits from comparing distorted images to pristine ones, while maintaining the practical advantages of an NR-IQA model.

Our work is more closely aligned with the earlier formulation of non-aligned reference IQA introduced by Liang et al.~\cite{liang2016image}. They proposed assessing image quality using a reference image from a similar scene that cannot be geometrically transformed to align with the target image. Their method applies affine transformations to generate non-aligned references, employing augmentation strategies to train a deep convolutional neural network on these inputs. Building on this foundation, our research focuses on the non-aligned reference IQA setting relevant to NVS-QA. Specifically, we explore whether image quality can be reliably assessed using reference views of nearby but non-aligned perspectives of the same scene. This setting reflects realistic use-cases in NVS-QA where some content overlap exists, but geometric and viewpoint disparities preclude pixel-level alignments. We posit that this scenario offers a more applicable and practical framework for evaluating the perceptual quality of synthesized novel views. The results we present later further corroborate the initial assumption.

\subsection{Novel View Synthesis IQA}

A recent study found that the visual degradations in NeRF-generated views differ significantly from those in traditional IQA datasets, causing existing IQA models trained on other types of synthetic and real algorithmic distortions to be less effective when evaluating synthesized novel views~\cite{zhang2025evaluating}. This early study primarily focused on front-facing camera trajectories, limiting their applicability to full 3D scene reconstructions. More recent approaches have extended their evaluations to include both front-facing and 360$^\circ$ views, better aligning with the use cases of modern novel view synthesis (NVS) techniques~\cite{martin2025gs}. However, these models often employ Pre-rendered Video Sequences (PVS) and rely on subjective video quality assessments, as seen in ~\cite{explicitnqa, pedronvs,tabassum2024quality}. These datasets and methods assume access to temporal information, rendering them unsuitable for evaluating quality metrics like ours, which operate purely in the spatial domain without leveraging temporal cues. Nonetheless, such PVS-based datasets remain valuable for the task of video quality assessment in NVS.

The advent of Gaussian Splatting (GS)~\cite{kerbl20233d}, a newer 3D scene representation technique based on superimposed anisotropic 3D Gaussians, has led to leaps in rendering efficiency and visual fidelity. GS offers a compelling alternative to NeRFs~\cite{mildenhall2021nerf}, enabling real-time rendering while maintaining high image quality. However, existing NVS-QA datasets do not yet capture the full range of degradations introduced by GS pipelines, making it essential to revisit metric development and evaluation strategies in this context. Towards this, \cite{zhang2025evaluating} and \cite{martin2025gs} proposed new benchmarks, but they too adopt the PVS setting. By contrast, our work emphasizes evaluating distortions in rendered novel views of neighbouring reference views.

Earlier NR-IQA methods such as NIQSV\cite{niqsv} were among the first to target NVS content, but they predate the recent advancements in NeRFs and GS and hence fall short of addressing the new challenges posed by these models. More recent work by Qu et al.\cite{qunerfnqa} introduced an NR-IQA method specifically tailored for NeRF-based NVS and later enhanced it using self-supervised learning~\cite{qu2025nvs}. Their approach involves training on algorithmically distorted outputs of real NVS systems to simulate artifacts. By contrast, our method trains on synthetically generated distortions using a contrastive learning framework, aiming for broader generalization across unseen degradations.

\subsection{Contrastive Learning in IQA}

Recent works have shown that contrastive learning can effectively model image quality without relying on large labeled datasets. In CONTRIQUE~\cite{madhusudana2022image}, distortion type and severity are used as pseudo-labels in a self-supervised framework to learn embeddings that generalize well across synthetic and authentic distortions. ARNIQA\cite{agnolucci2024arniqa} extends this by aligning representations of differently sourced patches with identical degradations, focusing on learning a distortion manifold rather than content-specific features. SaTQA~\cite{shi2023transformer} uses a supervised contrastive approach, explicitly leveraging distortion labels to guide feature learning, which are later fused with perceptual features from a CNN-Transformer backbone. The high performance of these methods highlights the ability of contrastive learning to capture perceptual degradations in both NR and FR-IQA settings.

% Also mention dreamsim here and then explain the part below

% Using synthetic data makes our approach more generic as it does not require 
% our approach less dependent on .. cite papers on contrastive learning...

\section{Evaluating Synthesized Novel Views using Non-Aligned Reference Views}\label{sec:study}

% couple of lines about various data sources and that they are publicly available - reliable....
\subsection{Synthesizing Novel Views}\label{sec:synthesizing_data}

% We constructed our test set by training three NeRFs~\cite{mildenhall2021nerf,muller2022instant,chen2022tensorf} and two Gaussian Splatting models~\cite{kerbl20233d,xu2024splatfacto} across 17 scenes from the NeRFStudio dataset~\cite{nerfstudio}. To evaluate the robustness of perceptual metrics under different data regimes, we utilized four train-test split configurations.
% : 90-10, 80-20, 70-30, and 50-50. 

\begin{figure}[t]
    \centering
    \includegraphics[width=0.9\linewidth]{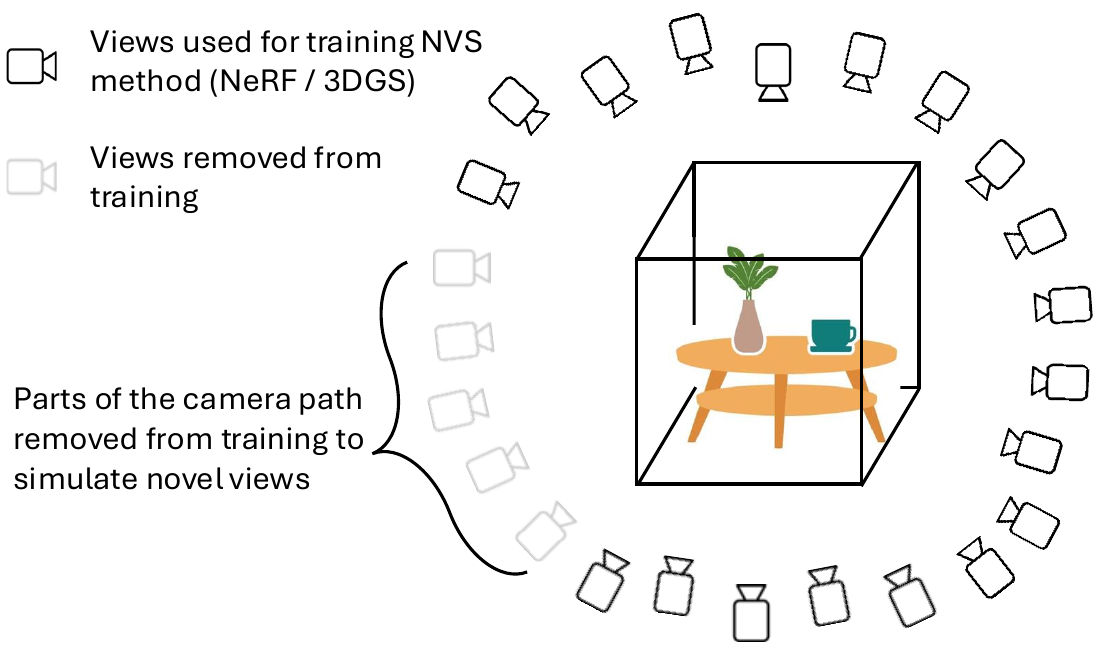}
    \caption{Views along the camera trajectory used to train the NeRF/GS models are indicated, while the excluded segments simulate novel views. These segments are used to create distorted image triplets with both aligned and non-aligned references.}
    \vspace{-0.1in}
    \label{fig:test_views}
\end{figure}

We constructed our test set by training three NeRFs~\cite{mildenhall2021nerf,muller2022instant,chen2022tensorf} and two Gaussian Splatting models~\cite{kerbl20233d,xu2024splatfacto} across 17 scenes from the NeRFStudio dataset~\cite{nerfstudio}. To evaluate the robustness of perceptual metrics under varying data regimes, we employed four train-test split configurations (90-10, 80-20, 70-30, and 50-50). Training 5 different NeRF/GS methods across 17 scenes with 4 train-test split settings results in a total of 340 trained models. Additionally, the selection of distinct camera trajectory segments, as illustrated in Figure~\ref{fig:test_views}, allows us to replicate degradations commonly encountered in novel views. For each triplet, originally with an aligned reference, we generated additional triplets by replacing the reference with a neighbouring, non-aligned view extracted from the same source video. These variations, applied across all train-test configurations, yielded over \textit{500,000 unique triplets}. We evaluated these triplets using the DreamSim model~\cite{fu2024dreamsim} to assess whether the relative ranking between the two distorted images—each synthesized by different models (NeRF/GS model1 vs. NeRF/GS model2)—changes when using a non-aligned reference instead of the original aligned one. This evaluation revealed that, depending on the scene, anywhere between 24,000 and 50,000 triplets exhibited a flipped ranking under the new reference condition.

% To evaluate the robustness of perceptual metrics under different data regimes, we utilized four train-test split configurations (90-10, 80-20, 70-30, and 50-50.). Training 5 different methods (NeRF/GS) on 17 scenes using 4 different train-test setups results in 340 trained models. Moreover, the use of diverse train-test splits, combined with the selection of varying segments of camera trajectories across scenes (as shown in Figure~\ref{fig:test_views}), enables us to replicate the types of degradations commonly observed in novel views. 

To ensure diversity among the selected samples, we employed Bayesian Optimization to select triplets exhibiting high variance across a wide range of test views (to capture scene diversity), train-test splits (to vary the degree of degradation), the distance between the nearby non-aligned reference view and the original aligned view (to account for diversity in content overlap reference viewpoints), and NeRF/GS methods (to account for different types of degradation). From each scene, we selected randomly 200 diverse samples, resulting in an initial pool of 3,400 triplets. A subsequent visual sanity check led to the removal of approximately 100 ambiguous triplets per scene (those with significant disagreement among human raters), yielding a refined set of 1,750 triplets.

\subsection{User Study}
\label{sec:user_study}

We then gathered perceptual ratings from five expert annotators having extensive expertise in the field of image and video quality assessment. They were asked to indicate which of the two distorted images exhibited higher visual quality relative to a nearby, non-aligned reference. Triplets with strong inter-rater disagreement were excluded, and each remaining triplet was assigned a binary label. \textit{Our final test dataset consists of 1,035 triplets, exhibiting high diversity in scene content, types of degradation, and levels of degradation.}

% : 0 if the majority favored \texttt{Image\_0}, and 1 if \texttt{Image\_1} was preferred.

% \textcolor{red}{Following the ratings (ADD SOME VALUE), Triplets with strong inter-rater disagreement were further excluded.} Each remaining triplet was assigned a binary label: 0 if the majority favored \texttt{Image\_0}, and 1 if \texttt{Image\_1} was preferred.

% \textit{Our final test dataset consists of 1,035 triplets, exhibiting high diversity in scene content, types of degradation, and levels of degradation.} %Based on the results from our benchmark in Table~\ref{tab:bench}, our metric showcases high performance on both aligned and non-aligned reference. We do not expose our metric to any NVS data and it performs well purely trained on synthetic dataset using contrastive learning which we explain the next section

\subsection{Training Dataset}\label{sec:dataset}
\begin{figure}[t]
  \centering
  \includegraphics[width=0.9\columnwidth]{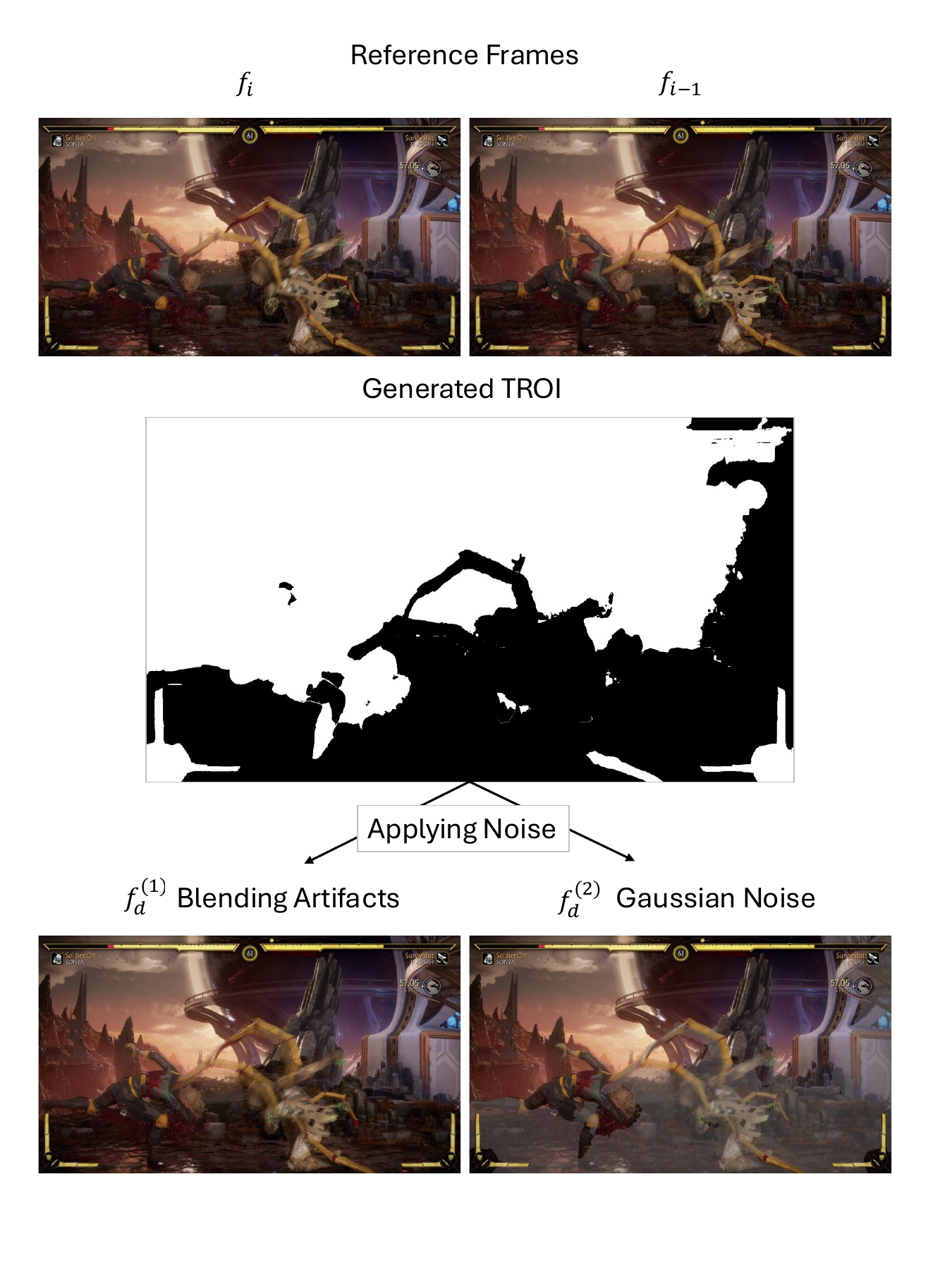}
  \caption{An example illustrating the synthetic dataset generation strategy, where distortions are applied locally within Temporal Regions of Interest (TROIs).}
  \label{fig:synthetic-distorted}
  \vspace{-0.2in}
\end{figure}

A major bottleneck in training models for NVS quality assessment is the lack of accessible and appropriate training data. To address this, we constructed a large-scale, high-resolution dataset specifically designed to support the development of robust and generalizable models. The dataset was sourced from over 120 lossless 4K video recordings, spanning a wide range of content types, including gaming, and frame rates (30fps and 60fps) to reflect the diversity of real-world rendering conditions. Our training dataset contains no videos or frames generated by NVS methods, and no data was sourced from existing NVS datasets, in order to avoid potential biases—both in terms of content and distortion types. Rather than splitting this dataset into training and testing subsets—which could risk overfitting to dataset-specific distortions—we opted to create a synthetic dataset that better aligns with the types of artifacts commonly seen in NVS outputs. While previous efforts such as the KADID dataset and the ARNIQA model have introduced distortions uniformly across entire images, our approach applies distortions selectively to different spatial regions. This more accurately mimics the localized and non-uniform artifacts characteristic of NVS models, enabling the model to learn more realistic and transferable quality degradation patterns.

The dataset was created with multiple synthetic distortions similar to the KADID dataset, but with a key difference: distortions were not applied across the entire frame. Instead, they were limited to areas exhibiting the most temporal change, which we refer to throughout this paper as Temporal Regions of Interest (TROI). A TROI is represented by a binary mask that defines where the distortion is applied. To compute the TROI, we used RAFT~\cite{teed2020raft} to estimate the optical flow between the current and previous frames. The resulting optical flow vectors were thresholded to define motion-based ROIs—localized areas most likely to reveal synthesis-related artifacts. This motion-aware ROI generation strategy enabled the dataset to be focused on perceptually critical regions in the NVS task, where distortions were both more common and more impactful on perceived quality. We applied a range of motion thresholds from 30\% to 85\%, while allowing distortions to appear over varying portions of the frame. In total, we introduced over 32 types of synthetic distortions, each applied at five intensity levels. These distortions were blended into the original frames only within the defined TROIs, ensuring that global image structures remained intact and visually plausible outside of the artifact-affected areas. This design provided a wide spectrum of realistic degradations while preserving the overall spatial consistency of the data

From the dataset, we constructed three triplets per sample centered on a target frame \( f_i \). To select the nearby reference, we randomly selected a nearby frame \( f_{i \pm k} \), where \( k \sim \mathcal{U}(1, 15) \), provided no scene change occurs. This setup enabled the model to learn from temporally unaligned references at varying distances from the target frame.

\section{Model Training}\label{sec:model}

\begin{figure}[t]
  \centering
  \includegraphics[width=0.96\columnwidth]{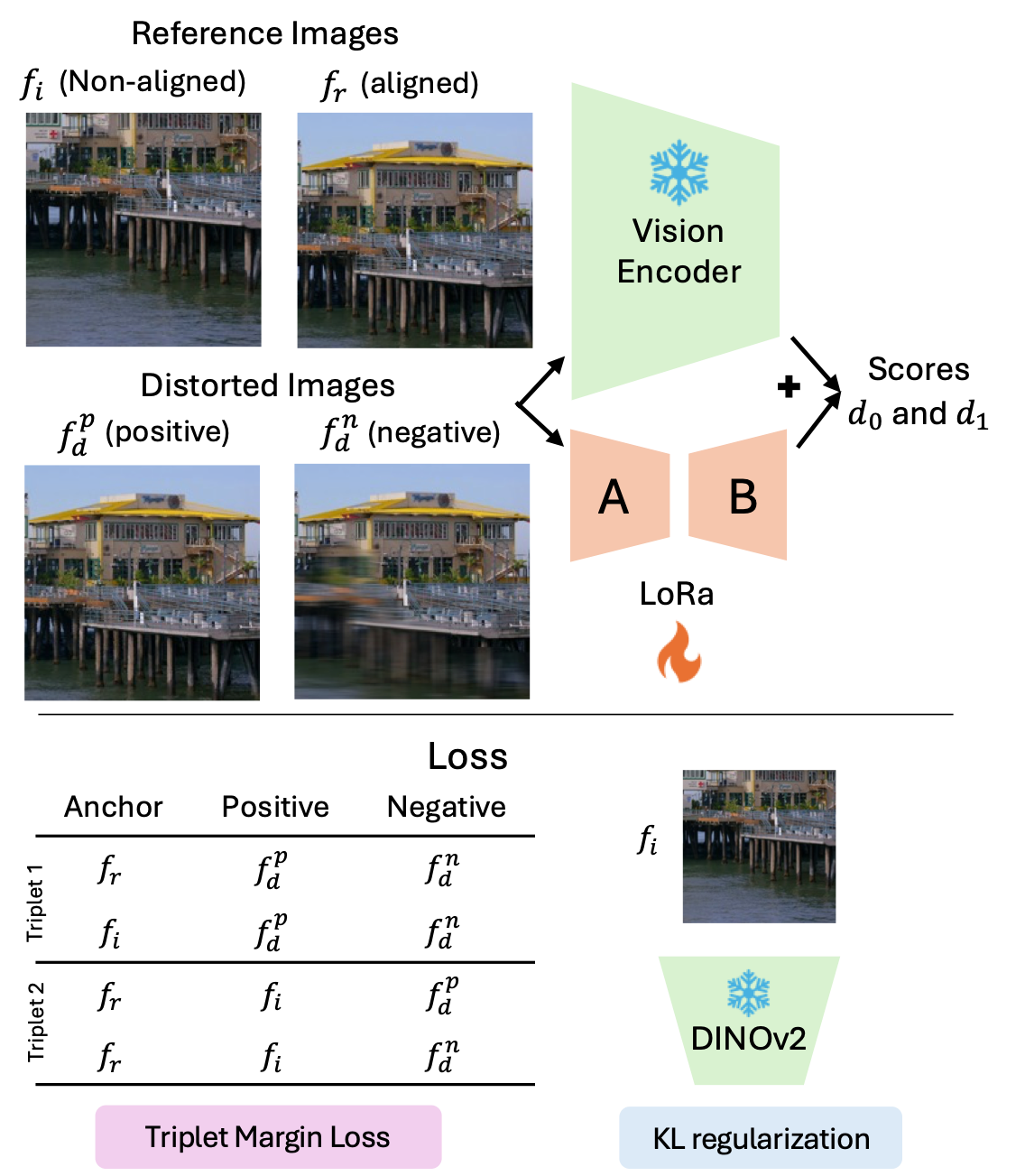}
  \caption{Overview of our model architecture. The network consists of a LoRA-enhanced DINOv2 backbone with dual outputs for embedding, trained using contrastive and auxiliary losses.}
  \label{fig:model-architecture}
  \vspace{-0.2in}
\end{figure}

Inspired by work on perceptual similarity~\cite{fu2024dreamsim}, we applied LoRA fine-tuning to the DINOv2 foundation model, extending it with a contrastive triplet loss and a KL divergence regularizer. This encourages a robust, human-aligned representation that remains close to the original DINOv2 space—preserving generalization while adapting to the specific demands of NVS quality assessment.
Figure \ref{fig:model-architecture} illustrates the training setup and model design.

\subsection{Contrastive Learning}

For each sample, we construct a training triplet centered on a target frame \( f_i \). A reference frame \( f_r \) is randomly selected within a window of \(\pm 15\) frames from \( f_i \), and two distorted versions of \( f_i \), denoted as \( f_d^{p} \) and \( f_d^{n} \), are generated, where \( f_d^{n} \) contains more severe artifacts than \( f_d^{p} \).

We use two cosine-distance-based triplet losses:
\begin{itemize}[topsep=0pt, partopsep=0pt, itemsep=0pt, parsep=0pt]
    \item \textbf{Triplet 1}: The anchor is either \( f_i \) or \( f_r \); the positive is \( f_d^{p} \); and the negative is \( f_d^{n} \). This triplet uses a margin \( m = 0.3 \).
    \item \textbf{Triplet 2}: The anchor is \( f_r \); the positive is \( f_i \); and the negative is either \( f_d^{p} \) or \( f_d^{n} \), randomly selected. This triplet uses a smaller margin \( m = 0.1 \).
\end{itemize}
We define the triplet margin loss as:
\[
\mathcal{L}_{\text{triplet}} = \max\left(0, d(\mathcal{E}(a), \mathcal{E}(n)) - d(\mathcal{E}(a), \mathcal{E}(p)) + m\right),
\]
where \( \mathcal{E}(\cdot) \) denotes the learned embedding, and \( a \), \( p \), and \( n \) are the anchor, positive, and negative samples respectively. The two triplet terms are computed independently and weighted equally in the total loss.

\subsection{IQA Model Supervision}\label{sec:iqa_supervision}
Not all triplets are consistent with human judgments of perceptual quality, which refers to determining which of two distorted images more closely resembles the reference in terms of perceived visual fidelity. As a result, some positive anchors may not represent true positives in the context of perceptual quality assessment. Following the supervised triplet formulation introduced by Wang et al.\cite{wang2024crossscore}, we assigned supervision labels to triplets involving an aligned reference and used the same labels for the corresponding non-aligned pairs. This scheme encouraged the model to assign similar perceptual distance scores regardless of whether the reference was aligned or non-aligned.

For supervision, we utilized DISTS~\cite{Ding20} and DeepDC~\cite{zhu2023DeepDC}, which exhibit strong alignment with human judgments on aligned-reference samples (see Table~\ref{tab:bench}). To ensure label reliability, we discarded triplets where the DISTS and DeepDC scores between the two distorted images were too close, suggesting ambiguity. For the remaining triplets, we removed any samples where the DeepDC predicted score disagrees with DISTS in terms of triplet ranking. This dual-stage filtering ensured that only triplets having consistent and confident supervision were retained, resulting in a curated set of 63,000 high-confidence examples to be used for contrastive training. We refer to this setup as `IQA model supervision,' with results detailed in Table~\ref{tab:ablation}.

% \textbf{While this automated supervision enables scalable training, it may introduce labelling errors due to metric limitations. Incorporating human judgements could improve label accuracy but was not feasible for large-scale dataset construction.}

\subsection{KL Divergence Regularization}

To maintain alignment with the pretrained representation space of DINOv2, we introduce a KL divergence loss between the softmax-normalized embeddings of the target frame \( f_i \) and those from a frozen version of the backbone: \vspace{-0.1in}

\[
\mathcal{L}_{\text{KL}} = D_{\text{KL}}\left( \text{softmax}\left( \frac{\mathcal{E}(f_i)}{T} \right) \bigg\| \text{softmax}\left( \frac{\mathcal{E}_{\text{frozen}}(f_i)}{T} \right) \right),
\]

\noindent where \( \mathcal{E}(\cdot) \) and \( \mathcal{E}_{\text{frozen}}(\cdot) \) are the learned and frozen embeddings, respectively, and \( T \) is a temperature parameter.

To improve training stability and prevent early collapse, we adopted temperature annealing, gradually increasing \( T \) from 0.01 to 1.0, enabling the model to first align coarse similarities and progressively refine semantic alignment.

% over the course of training. This approach allows the model to focus initially on coarse similarity alignment and refine semantic alignment as training progresses.

% \subsection{Segmentation Loss}

% To further guide the model toward perceptually relevant regions, we train a segmentation head to predict binary masks representing Temporal Regions of Interest (TROI), indicating where distortions are most likely to occur. The segmentation output is supervised using a binary cross-entropy loss:
% \[
% \mathcal{L}_{\text{seg}} = \text{BCE}(\mathcal{S}(f_i), y),
% \]
% where \( \mathcal{S}(\cdot) \) is the segmentation head and \( y \) is the ground truth TROI mask.

\subsection{Total Loss}
%\lambda_{\text{seg}} \mathcal{L}_{\text{seg}} +
The final loss function is a weighted combination:\vspace{-0.05in}
\[
\mathcal{L}_{\text{total}} =  \lambda_1 \mathcal{L}_{\text{triplet1}} + \lambda_2 \mathcal{L}_{\text{triplet2}} + \lambda_{\text{KL}} \mathcal{L}_{\text{KL}},
\]
with weights empirically set to \( \lambda_1 = 1.0 \), \( \lambda_2 = 1.0 \), and \( \lambda_{\text{KL}} = 0.05 \).

\section{Experiments and Results}
\label{sec:exp}

\subsection{Training Details}

We initialized the model with a LoRA-augmented DINOv2 backbone and used a frozen version of the same model to compute the KL loss. All models were trained using the AdamW optimizer with a learning rate of \(1 \times 10^{-5}\), batch size 16, and input resolution of \(518 \times 518\). Training was performed over 80 epochs using our synthetic TROI-annotated dataset, as described in Section~\ref{sec:dataset}.

\subsection{Inference and Evaluation}

% At inference time, we measure the perceptual similarity between a processed (e.g., synthesized) frame and a reference frame—either the exact ground truth frame \( f_i \) or a temporally nearby frame \( f_r \)—by computing the cosine similarity between their embeddings in the model's latent space. This similarity serves as a proxy for visual quality, enabling direct comparisons between candidate outputs in scenarios with either aligned or non-aligned references.

At inference time, we measure the perceptual similarity between a synthesized frame and its reference—either the aligned ground truth \( f_r \) or a non-aligned frame \( f_i \)—by computing the cosine similarity between their embeddings in the model's latent space. This similarity serves as a proxy for visual quality, enabling direct comparisons between synthesized views using either aligned or non-aligned references.

To ensure fair and robust evaluation, we compute accuracy over the final 10 epochs of training, using the non-aligned NVS test set introduced in Section~\ref{sec:synthesizing_data}. Accuracy reflects the agreement between a model's predicted preference and the human-annotated label, aggregated over all pairwise comparisons.

\subsection{Quantitative Results}
\subsubsection{NVS NAR-IQA Benchmark}

\begin{figure*}[t]
  \centering
  \includegraphics[width=\textwidth]{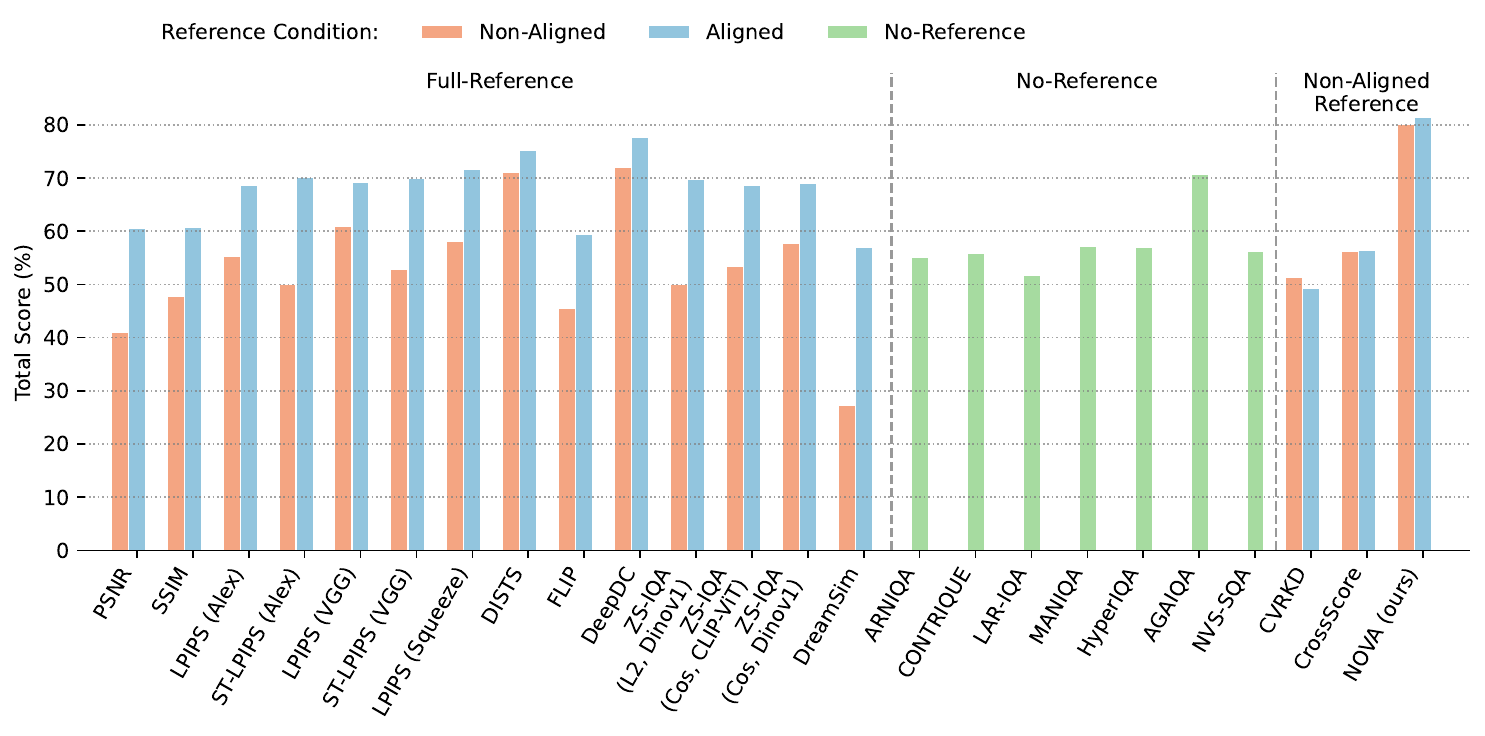}
  \vspace{-0.3in}
  \caption{Quantitative comparison (accuracy in \%) of IQA models on the NVS NAR-IQA benchmark. We evaluate full-reference, non-aligned reference, and no-reference metrics under both aligned and non-aligned reference conditions.}
  \label{fig:bench}
  \vspace{-0.1in}
\end{figure*}

% Figure~\ref{tab:bench} (also Table~\ref{tab:bench} in supplementary) summarizes the performance of our proposed model against a wide range of IQA baselines—including full-reference (FR), no-reference (NR), and non-aligned reference (NAR) approaches—on our challenging NVS NAR-IQA benchmark. 
Figure~\ref{fig:bench} (with detailed results provided in Table~\ref{tab:bench} of the supplementary material) presents the performance of our proposed model against a wide range of IQA baselines—including full-reference (FR), no-reference (NR), and non-aligned reference (NAR) approaches—on our challenging NVS NAR-IQA benchmark. This test set comprises 1,035 triplets across 17 diverse scenes, selected to capture a wide range of degradation types, severity levels, and content overlap between the reference and distorted views. As a result, each sample in the dataset is unique in terms of scene characteristics and distortion attributes.

% , encompassing full-reference (FR), no-reference (NR), and non-aligned reference (NAR) approaches, on the challenging NVS NAR-IQA benchmark. 

% Each sample features a unique configuration of reference-distortion overlap and scene complexity.

% This dataset comprises 17 diverse novel view synthesis (NVS) scenes, with careful attention paid to diversity in degradation severity, content overlap between the reference and distorted views, and degradation types. The methodology employed to ensure this diversity is described in Section~\ref{sec:synthesizing_data}. As a result, each sample in the dataset is unique with respect to scene characteristics and distortion attributes.

Our model achieved state-of-the-art results in both aligned and non-aligned reference settings. Specifically, it reaches a mean accuracy of \textbf{81.45\%} for aligned references and \textbf{80.19\%} for non-aligned references, outperforming all other evaluated models, including the pretrained DINOv2 backbone and the prior NAR-IQA models. When reporting average accuracy over the final 10 epochs, we observed a mean of \textbf{79.21\%} with a standard deviation of \textbf{0.49\%}, demonstrating the stability and robustness of our approach.

% Importantly, our model reduces the performance gap between aligned and non-aligned settings to just \textbf{1.26 percentage points}, whereas traditional FR-IQA metrics such as LPIPS (VGG) and DeepDC exhibit much sharper declines (e.g., 69.2\% to 61.0\%, and 77.7\% to 72.1\%, respectively).
% This highlights our model's ability to retain perceptual discrimination even when spatial alignment is degraded, an essential property for practical NVS evaluation.

Our model's performance gap between the aligned and the non-aligned settings is only \textbf{1.26\%}, whereas all other models exhibit much sharper declines. This underscores our model’s robustness in perceptual evaluation under spatial misalignment, a key requirement for effective NVS assessment. Please note that DreamSim's low performance is partly due to our selection of scenes where its ranking flipped between aligned and non-aligned settings, as discussed in Section~\ref{sec:synthesizing_data}.

NR-IQA methods, while alignment-agnostic by design, underperformed on this benchmark, likely due to their limited ability to learn NVS types of artifacts. Likewise, the self-supervised NAR-IQA model CVRKD also underperformed, which may be due to a sensitivity to the distribution of high- and low-quality image pairs in its training regime. 

Taken together, these results highlight the efficacy of our localized distortion training pipeline in capturing perceptually meaningful features, enabling generalization across unseen distortions, scene layouts, and viewpoint shifts.

\subsubsection{NVS-QA Benchmark}

We evaluated our proposed NVS-IQA model NOVA on the NVS-QA benchmark~\cite{pedronvs}. In the non-aligned reference setting, we selected a frame from the video that is either 5 time steps before the ground-truth aligned reference or, if unavailable, 5 steps after. This setup simulates misalignment encountered in real-world NVS use cases.

% We evaluate our proposed metric, NAR-NVSQA, on the NVS-QA benchmark~\cite{pedronvs}, which we include both aligned and unaligned references for view synthesis quality assessment. The \textit{unaligned reference} is generated by selecting the frame 5 steps before the ground-truth reference, or 5 steps after if the earlier frame is unavailable. 

Table~\ref{tab:metric_correlation} presents the Pearson (PLCC) and Spearman (SRCC) correlation coefficients between DMOS and various quality metrics. Our proposed method achieves the highest SRCC among all IQA models, with a PLCC of -0.73 and SRCC of -0.83 in the aligned setting, and a PLCC of -0.73 and SRCC of -0.78 with non-aligned references. Although DISTS yields a slightly higher PLCC, we attribute this to its training paradigm, which directly optimizes for correlation by targeting MOS scores. By contrast, our model prioritizes rank consistency, as evidenced by its higher SRCC. Our method outperforms all other metrics on this benchmark, including video-based quality metrics. 

% A more detailed discussion of this distinction is included in the supplementary material.

% Table~\ref{tab:metric_correlation} reports the Pearson (PLCC) and Spearman (SRCC) correlation coefficients between DMOS and various quality metrics. Our proposed method outperforms all IQA models in terms of SRCC, achieving PLCC = -0.73, SRCC = -0.83 in the aligned setting and PLCC = -0.73, SRCC = -0.78 with non-aligned references. While DISTS exhibits a slightly higher PLCC, we attribute this to its training paradigm, which optimizes for correlation by targeting MOS scores. In contrast, our model is optimized for rank consistency, as reflected by the higher SRCC. A detailed discussion of this distinction is provided in the supplementary material.

Unlike conventional IQA models, our proposed NOVA model is explicitly trained to handle non-aligned references using contrastive supervision and synthetic distortions. Its strong performance across both settings, despite spatial misalignment, demonstrates its robustness and practical applicability to real-world NVS tasks, where aligned ground truth is rarely available.

\begin{table}[t]
\setlength{\tabcolsep}{3pt}
\centering
\begin{tabular}{lcccc}
\toprule
\textbf{Metric} & \multicolumn{2}{c}{\textbf{Aligned}} & \multicolumn{2}{c}{\textbf{Non-aligned}} \\
\cmidrule(lr){2-3} \cmidrule(lr){4-5}
 & \textbf{PLCC} & \textbf{SRCC} & \textbf{PLCC} & \textbf{SRCC} \\
\midrule
LPIPS (Alex)     & -0.63 & -0.79 & -0.51 & -0.52 \\
LPIPS (Sqz)      & -0.65 & -0.80 & -0.57 & -0.56 \\
LPIPS (VGG)      & -0.65 & -0.72 & -0.59 & -0.60 \\
ST-LPIPS (Alex)     & -0.58 & -0.74 & -0.43 & -0.45 \\
ST-LPIPS (VGG)     & -0.61 & -0.75 & -0.49 & -0.51 
\\
DreamSim            & -0.58 & -0.80& -0.54 & 0.59
\\
DISTS            & \textbf{-0.75} & -0.71 & -0.67 & -0.65 \\
DeepDC           & -0.69 & -0.82 & -0.61 & -0.58 \\
DINOv2           & -0.62 & -0.81 & -0.61 & -0.70 \\
\textbf{NOVA (ours)} & -0.73 & \textbf{-0.83} & \textbf{-0.73} & \textbf{-0.78} \\
\bottomrule
\end{tabular}
\vspace{-0.1in}
\caption{Correlation between IQA models and DMOS using PLCC and SRCC under aligned and unaligned references.}
\label{tab:metric_correlation}
\vspace{-0.05in}
\end{table}

\begin{figure}[t]
  \centering
  \includegraphics[width=\linewidth]{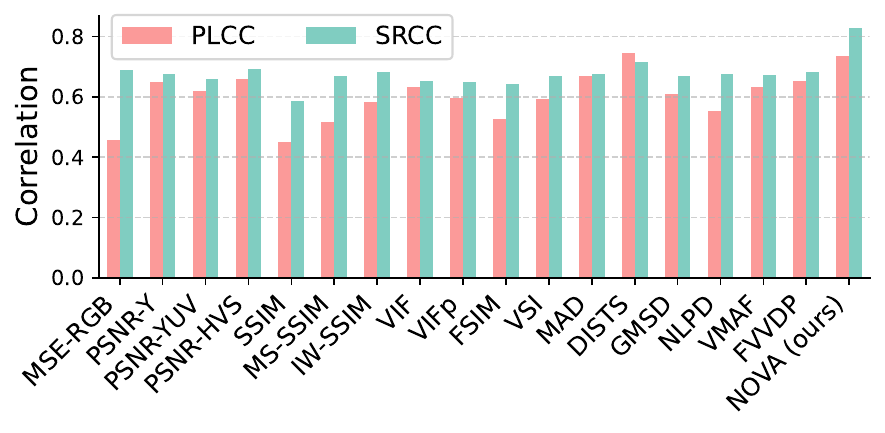}
  \vspace{-0.3in}
  \caption{Correlation between objective metrics and DMOS using PLCC and SRCC on NVS-QA dataset.}
  \label{fig:metric_correlationv2}
  \vspace{-0.2in}
\end{figure}

As the NVS-QA dataset comprises Pre-rendered Video Sequences (PVS), we complement the above evaluation with additional correlation analyses between DMOS and objective quality metrics under the PVS setting with aligned references. Figure~\ref{fig:metric_correlationv2} reports PLCC and SRCC for classical and learned metrics—including PSNR variants, SSIM-based methods, and video-based quality metrics. 

% Results under non-aligned reference conditions are omitted, as most metrics—relying on pixelwise error or structural similarity—fail under geometric or semantic misalignment. Nonetheless, 

% metrics operating in perceptual embedding spaces (e.g., DreamSim, DeepDC) show greater resilience, highlighting the limitations of traditional approaches for NVS artifacts.

% \begin{table}[h]
% \centering
% \begin{tabular}{l cc}
% \toprule
% \textbf{Metric} & \textbf{PLCC} & \textbf{SRCC} \\
% \midrule
% MSE-RGB     & -0.4568 & -0.6888 \\
% PSNR-Y      &  0.6484 &  0.6754 \\
% PSNR-YUV    &  0.6199 &  0.6592 \\
% PSNR-HVS    &  0.6601 &  0.6932 \\
% SSIM        &  0.4508 &  0.5862 \\
% MS-SSIM     &  0.5158 &  0.6698 \\
% IW-SSIM     &  0.5812 &  0.6814 \\
% VIF         &  0.6329 &  0.6517 \\
% VIFp        &  0.5963 &  0.6478 \\
% FSIM        &  0.5253 &  0.6407 \\
% VSI         &  0.5914 &  0.6696 \\
% MAD         & -0.6675 & -0.6747 \\
% DISTS       & -0.7461 & -0.7151 \\
% GMSD        & -0.6098 & -0.6703 \\
% NLPD        & -0.5515 & -0.6764 \\
% VMAF        &  0.6325 &  0.6704 \\
% FVVDP       &  0.6528 &  0.6806 \\
% \textbf{NOVA (ours)} & \textbf{-0.7333} & \textbf{-0.8273} \\
% \bottomrule
% \end{tabular}
% \caption{Correlation between objective metrics and DMOS using PLCC and SRCC.
% }
% \label{tab:metric_correlationv2}
% \end{table}

\subsection{Ablation Study}

As demonstrated in the ablation study in Table~\ref{tab:ablation}, the combination of IQA model supervision, TROI, triplet loss, and KL divergence loss contributes to observed performance improvements. IQA supervision facilitates learning from contrastive pairs, which are determined through triplet combinations used in the triplet loss. Meanwhile, the KL divergence loss serves as a regularizer, leveraging DINOv2’s capabilities to evaluate perceptual quality.

\section{Summary}
\label{sec:summary}
In this paper we present the first study of non-aligned reference image quality assessment (NAR-IQA) for novel view synthesis (NVS). Existing IQA methods were typically formulated for strictly aligned reference–distorted pairs and therefore face inherent limitations when applied to NVS, where larger spatial misalignment is common. To quantify and benchmark this challenge, we construct a comprehensive dataset of 17 scenes generated with NeRF and Gaussian Splatting models, encompassing a broad spectrum of distortion types, severities, and reference–distorted image content overlap (Section~\ref{sec:study}). We develop a new NAR-IQA model for NVS, NOVA, which is trained on a large synthetic dataset using contrastive learning and IQA model supervision, with a DINOv2 backbone (Section~\ref{sec:model}). NOVA achieves state-of-the-art performance on the proposed benchmark (Figure~\ref{fig:bench}). It further outperforms classical, learned, and video-based metrics on the NVS-QA dataset, demonstrating strong correlation with human judgments (Table~\ref{tab:metric_correlation} and Figure~\ref{fig:metric_correlationv2}).

% \begin{table}[t]
% \centering
% \begin{tabular}{l c c c c c c}
% \toprule
%  & & \multicolumn{3}{c}{Loss} & \multicolumn{2}{c}{Reference Type} \\
% \textbf{Model} & \rotatebox{90}{\shortstack{IQA model \\ Supervision}} & \rotatebox{90}{Hinge} & \rotatebox{90}{Triplet} & \rotatebox{90}{KL} & \rotatebox{90}{Aligned} & \rotatebox{90}{\shortstack{Non-\\Aligned}} \\
% \midrule
% DINOv2 &  &  &  &  & 74.49 & 71.79 \\
% + LORA &  &  & \checkmark &  & 66.18 & 62.61 \\
% + LORA & \checkmark & \checkmark &  &  & 76.43 & 75.94 \\
% + LORA & \checkmark &  & \checkmark &  & 78.45 & 77.58 \\
% + LORA & \checkmark &  & \checkmark & \checkmark & 81.45 & 80.19 \\
% \bottomrule
% \end{tabular}
% \vspace{-0.1in}
% \caption{Ablation Study}
% \label{tab:ablation}
% \vspace{-0.2in}
% \end{table}

\begin{table}[t]
\centering
\begin{tabular}{l c c c c c c c}
\toprule
 & & & \multicolumn{3}{c}{Loss} & \multicolumn{2}{c}{Reference Type} \\
 \cmidrule(lr){4-6} \cmidrule(lr){7-8}
\textbf{Model} & \rotatebox{90}{\shortstack{IQA model \\ Supervision}} &\rotatebox{90}{TROI} & \rotatebox{90}{Hinge} & \rotatebox{90}{Triplet} & \rotatebox{90}{KL} & \rotatebox{90}{Aligned} & \rotatebox{90}{\shortstack{Non-\\Aligned}} \\
\midrule
DINOv2 & &  &  &  &  & 74.49 & 71.79 \\
+ LORA & &\checkmark  &  & \checkmark &  & 66.18 & 62.61 \\
\hdashline
+ LORA & \checkmark &\checkmark & \checkmark &  &  & 76.43 & 75.94 \\
+ LORA & \checkmark &\checkmark &  & \checkmark &  & 78.45 & 77.58 \\
\hdashline
+ LORA & \checkmark & &  & \checkmark & \checkmark &76.04 &75.18 \\
+ LORA & \checkmark &\checkmark &  & \checkmark & \checkmark & 81.45 & 80.19 \\
\bottomrule
\end{tabular}
\vspace{-0.1in}
\caption{Ablation Study}
\label{tab:ablation}
\vspace{-0.2in}
\end{table}

\section{Limitations}
\label{sec:lim}

% We tried to simulate the novel views but in the future work we will need to generate novel views

% Work on video with novel and also work on flicker artifacts and also test for denoising methods like Nerflix and NeRF-Gan 

Although automated label collection using IQA models for supervision, as detailed in Section~\ref{sec:iqa_supervision}, enables efficient training, it may introduce labeling errors due to their imperfect alignment with human perception. While human judgments could improve label accuracy, large-scale annotation in this context is quite challenging.

NVS methods often produce temporal artifacts such as flickering and ghosting, making video-based quality evaluation particularly relevant~\cite{ghildyal2024quality}, similar to the PVS setting used in prior work~\cite{pedronvs,qu2025nvs}. Although our approach outperformed existing video quality models in the PVS setting on the dataset from\cite{pedronvs}, despite not explicitly handling temporal artifacts, incorporating temporal modeling remains a key focus for future work. 

\section{Conclusion}

% We proposed a novel NAR-IQA test dataset and model for assessing the quality of novel view synthesis without requiring pixel-aligned references. By leveraging localized synthetic distortions and contrastive learning with LoRA-enhanced DINOv2, our model achieves state-of-the-art performance across diverse scenes and NVS distortions. 

We propose a novel NAR-IQA dataset and model for NVS-QA, called NOVA, that does not require pixel-level alignment. Through contrastive learning on localized distortions and LoRA-enhanced DINOv2 features, our model achieved state-of-the-art performance across diverse settings. It significantly narrows the gap between aligned and non-aligned reference settings and shows strong correlation with human judgments, offering a practical solution for real-world NVS-QA. We plan to release our test dataset and model to support future research in this area.

\label{sec:conc}

{\small
\bibliographystyle{ieee_fullname}
\bibliography{egbib}

\begin{thebibliography}{10}\itemsep=-1pt

\bibitem{agnolucci2024arniqa}
Lorenzo Agnolucci, Leonardo Galteri, Marco Bertini, and Alberto Del~Bimbo.
\newblock {ARNIQA}: Learning distortion manifold for image quality assessment.
\newblock In {\em Proceedings of the IEEE/CVF Winter Conference on Applications of Computer Vision}, pages 189--198, 2024.

\bibitem{andersson2020flip}
Pontus Andersson, Jim Nilsson, Tomas Akenine-M{\"o}ller, Magnus Oskarsson, Kalle {\AA}str{\"o}m, and Mark~D Fairchild.
\newblock Flip: A difference evaluator for alternating images.
\newblock {\em Proceedings of the ACM on Computer Graphics and Interactive Techniques}, 3(2):15--1, 2020.

\bibitem{sangnie2020}
Sangnie Bhardwaj, Ian Fischer, Johannes Ball{\'e}, and Troy Chinen.
\newblock An unsupervised information-theoretic perceptual quality metric.
\newblock {\em Advances in Neural Information Processing Systems}, 33:13--24, 2020.

\bibitem{chen2022tensorf}
Anpei Chen, Zexiang Xu, Andreas Geiger, Jingyi Yu, and Hao Su.
\newblock Tensorf: Tensorial radiance fields.
\newblock In {\em European conference on computer vision}, pages 333--350, 2022.

\bibitem{Ding20}
Keyan Ding, Kede Ma, Shiqi Wang, and Eero~P. Simoncelli.
\newblock Image quality assessment: Unifying structure and texture similarity.
\newblock {\em IEEE Transactions on Pattern Analysis and Machine Intelligence}, pages 1--1, 2020.

\bibitem{fu2024dreamsim}
Stephanie Fu, Netanel Tamir, Shobhita Sundaram, et~al.
\newblock Dreamsim: Learning new dimensions of human visual similarity using synthetic data.
\newblock {\em Advances in Neural Information Processing Systems}, 36, 2024.

\bibitem{ghildyal2025foundation}
Abhijay Ghildyal, Nabajeet Barman, and Saman Zadtootaghaj.
\newblock Foundation models boost low-level perceptual similarity metrics.
\newblock In {\em IEEE International Conference on Acoustics, Speech and Signal Processing}, pages 1--5, 2025.

\bibitem{ghildyal2024quality}
Abhijay Ghildyal, Yuanhan Chen, Saman Zadtootaghaj, Nabajeet Barman, and Alan~C Bovik.
\newblock Quality prediction of ai generated images and videos: Emerging trends and opportunities.
\newblock {\em arXiv:2410.08534}, 2024.

\bibitem{ghildyal2022stlpips}
Abhijay Ghildyal and Feng Liu.
\newblock Shift-tolerant perceptual similarity metric.
\newblock In {\em European Conference on Computer Vision}, pages 91--107, 2022.

\bibitem{ghildyal2023attacking}
Abhijay Ghildyal and Feng Liu.
\newblock Attacking perceptual similarity metrics.
\newblock {\em Transactions on Machine Learning Research}, 2023.
\newblock Featured Certification.

\bibitem{jamshidi2025lar}
Nasim Jamshidi~Avanaki, Abhijay Ghildyal, Nabajeet Barman, and Saman Zadtootaghaj.
\newblock Lar-iqa: A lightweight, accurate, and robust no-reference image quality assessment model.
\newblock In {\em European Conference on Computer Vision Workshops}, pages 328--345, 2024.

\bibitem{kerbl20233d}
Bernhard Kerbl, Georgios Kopanas, Thomas Leimk{\"u}hler, and George Drettakis.
\newblock 3d gaussian splatting for real-time radiance field rendering.
\newblock {\em ACM Transactions on Graphics}, 42(4):139--1, 2023.

\bibitem{li2009reduced}
Qiang Li and Zhou Wang.
\newblock Reduced-reference image quality assessment using divisive normalization-based image representation.
\newblock {\em IEEE journal of selected topics in signal processing}, 3(2):202--211, 2009.

\bibitem{li2023less}
Xudong Li, Jingyuan Zheng, Xiawu Zheng, Runze Hu, Enwei Zhang, Yuting Gao, Yunhang Shen, Ke Li, Yutao Liu, Pingyang Dai, et~al.
\newblock Less is more: Learning reference knowledge using no-reference image quality assessment.
\newblock {\em arXiv:2312.00591}, 2023.

\bibitem{liang2016image}
Yudong Liang, Jinjun Wang, Xingyu Wan, Yihong Gong, and Nanning Zheng.
\newblock Image quality assessment using similar scene as reference.
\newblock In {\em European Conference on Computer Vision}, pages 3--18, 2016.

\bibitem{madhusudana2022image}
Pavan~C Madhusudana, Neil Birkbeck, Yilin Wang, Balu Adsumilli, and Alan~C Bovik.
\newblock Image quality assessment using contrastive learning.
\newblock {\em IEEE Transactions on Image Processing}, 31:4149--4161, 2022.

\bibitem{pedronvs}
Pedro Martin, António Rodrigues, João Ascenso, and Maria Paula~Queluz.
\newblock Nerf view synthesis: Subjective quality assessment and objective metrics evaluation.
\newblock {\em IEEE Access}, 13:26--41, 2025.

\bibitem{martin2025gs}
Pedro Martin, Ant{\'o}nio Rodrigues, Jo{\~a}o Ascenso, and Maria~Paula Queluz.
\newblock Gs-qa: Comprehensive quality assessment benchmark for gaussian splatting view synthesis.
\newblock {\em arXiv preprint arXiv:2502.13196}, 2025.

\bibitem{mildenhall2021nerf}
Ben Mildenhall, Pratul~P Srinivasan, Matthew Tancik, Jonathan~T Barron, Ravi Ramamoorthi, and Ren Ng.
\newblock Nerf: Representing scenes as neural radiance fields for view synthesis.
\newblock {\em Communications of the ACM}, 65(1):99--106, 2021.

\bibitem{muller2022instant}
Thomas M{\"u}ller, Alex Evans, Christoph Schied, and Alexander Keller.
\newblock Instant neural graphics primitives with a multiresolution hash encoding.
\newblock {\em ACM Transactions on Graphics}, 41(4):1--15, 2022.

\bibitem{prashnani2018pieapp}
Ekta Prashnani, Hong Cai, Yasamin Mostofi, and Pradeep Sen.
\newblock Pieapp: Perceptual image-error assessment through pairwise preference.
\newblock In {\em IEEE Conference on Computer Vision and Pattern Recognition}, pages 1808--1817, 2018.

\bibitem{qunerfnqa}
Qiang Qu, Hanxue Liang, Xiaoming Chen, Yuk~Ying Chung, and Yiran Shen.
\newblock Nerf-nqa: No-reference quality assessment for scenes generated by nerf and neural view synthesis methods.
\newblock {\em IEEE Transactions on Visualization and Computer Graphics}, 30(5):2129--2139, 2024.

\bibitem{qu2025nvs}
Qiang Qu, Yiran Shen, Xiaoming Chen, Yuk~Ying Chung, Weidong Cai, and Tongliang Liu.
\newblock Nvs-sqa: Exploring self-supervised quality representation learning for neurally synthesized scenes without references.
\newblock {\em arXiv:2501.06488}, 2025.

\bibitem{shi2023transformer}
Jinsnog Shi, Gao Pan, and Qin Jie.
\newblock Transformer-based no-reference image quality assessment via supervised contrastive learning.
\newblock In {\em AAAI}, 2024.

\bibitem{shi2023robust}
Wenbo Shi, Wenming Yang, and Qingmin Liao.
\newblock Robust content-variant reference image quality assessment via similar patch matching.
\newblock In {\em International Conference on Acoustics, Speech and Signal Processing}, pages 1--5. IEEE, 2023.

\bibitem{soundararajan2011rred}
Rajiv Soundararajan and Alan~C Bovik.
\newblock Rred indices: Reduced reference entropic differencing for image quality assessment.
\newblock {\em IEEE Transactions on Image Processing}, 21(2):517--526, 2011.

\bibitem{Su_2020_CVPR}
Shaolin Su, Qingsen Yan, Yu Zhu, Cheng Zhang, Xin Ge, Jinqiu Sun, and Yanning Zhang.
\newblock Blindly assess image quality in the wild guided by a self-adaptive hyper network.
\newblock In {\em IEEE/CVF Conference on Computer Vision and Pattern Recognition}, June 2020.

\bibitem{sun2022graphiqa}
Simeng Sun, Tao Yu, Jiahua Xu, Wei Zhou, and Zhibo Chen.
\newblock Graphiqa: Learning distortion graph representations for blind image quality assessment.
\newblock {\em IEEE Transactions on Multimedia}, 25:2912--2925, 2022.

\bibitem{sureddi2025triqa}
Rajesh Sureddi, Saman Zadtootaghaj, Nabajeet Barman, and Alan~C Bovik.
\newblock {TRIQA}: Image quality assessment by contrastive pretraining on ordered distortion triplets.
\newblock In {\em 2025 IEEE International Conference on Image Processing (ICIP)}, pages 1744--1749, 2025.

\bibitem{tabassum2024quality}
Shaira Tabassum and Seyed~Ali Amirshahi.
\newblock Quality of nerf changes with the viewing path an observer takes: A subjective quality assessment of real-time nerf model.
\newblock In {\em 2024 16th International Conference on Quality of Multimedia Experience (QoMEX)}, pages 88--91. IEEE, 2024.

\bibitem{nerfstudio}
Matthew Tancik, Ethan Weber, Evonne Ng, Ruilong Li, Brent Yi, Justin Kerr, Terrance Wang, Alexander Kristoffersen, Jake Austin, Kamyar Salahi, Abhik Ahuja, David McAllister, and Angjoo Kanazawa.
\newblock Nerfstudio: A modular framework for neural radiance field development.
\newblock In {\em ACM SIGGRAPH}, 2023.

\bibitem{teed2020raft}
Zachary Teed and Jia Deng.
\newblock Raft: Recurrent all-pairs field transforms for optical flow.
\newblock In {\em European Conference on Computer Vision}, pages 402--419, 2020.

\bibitem{niqsv}
Shishun Tian, Lu Zhang, Luce Morin, and Olivier Déforges.
\newblock Niqsv+: A no-reference synthesized view quality assessment metric.
\newblock {\em IEEE Transactions on Image Processing}, 27(4):1652--1664, 2018.

\bibitem{wang2024blind}
Huasheng Wang, Jiang Liu, Hongchen Tan, Jianxun Lou, Xiaochang Liu, Wei Zhou, and Hantao Liu.
\newblock Blind image quality assessment via adaptive graph attention.
\newblock {\em IEEE Transactions on Circuits and Systems for Video Technology}, 34(10):10299--10309, 2024.

\bibitem{wang2016reduced}
Shiqi Wang, Ke Gu, Xinfeng Zhang, Weisi Lin, Siwei Ma, and Wen Gao.
\newblock Reduced-reference quality assessment of screen content images.
\newblock {\em IEEE Transactions on Circuits and Systems for Video Technology}, 28(1):1--14, 2016.

\bibitem{wang2024crossscore}
Zirui Wang, Wenjing Bian, and Victor~Adrian Prisacariu.
\newblock Crossscore: Towards multi-view image evaluation and scoring.
\newblock In {\em European Conference on Computer Vision}, pages 492--510, 2024.

\bibitem{wang2004image}
Zhou Wang, Alan~C Bovik, Hamid~R Sheikh, and Eero~P Simoncelli.
\newblock Image quality assessment: from error visibility to structural similarity.
\newblock {\em IEEE transactions on Image Processing}, 13(4):600--612, 2004.

\bibitem{wang2005reduced}
Zhou Wang and Eero~P Simoncelli.
\newblock Reduced-reference image quality assessment using a wavelet-domain natural image statistic model.
\newblock In {\em Human vision and electronic imaging X}, volume 5666, pages 149--159. SPIE, 2005.

\bibitem{wang2003multiscale}
Zhou Wang, Eero~P Simoncelli, and Alan~C Bovik.
\newblock Multiscale structural similarity for image quality assessment.
\newblock In {\em The Thrity-Seventh Asilomar Conference on Signals, Systems \& Computers}, volume~2, pages 1398--1402. IEEE, 2003.

\bibitem{explicitnqa}
Yuke Xing, Qi Yang, Kaifa Yang, Yiling Xu, and Zhu Li.
\newblock Explicit-nerf-qa: A quality assessment database for explicit nerf model compression.
\newblock In {\em IEEE International Conference on Visual Communications and Image Processing}, pages 1--5, 2024.

\bibitem{xu2024splatfacto}
Congrong Xu, Justin Kerr, and Angjoo Kanazawa.
\newblock Splatfacto-w: A nerfstudio implementation of gaussian splatting for unconstrained photo collections.
\newblock {\em arXiv preprint arXiv:2407.12306}, 2024.

\bibitem{yang2022maniqa}
Sidi Yang, Tianhe Wu, Shuwei Shi, Shanshan Lao, Yuan Gong, Mingdeng Cao, Jiahao Wang, and Yujiu Yang.
\newblock Maniqa: Multi-dimension attention network for no-reference image quality assessment.
\newblock In {\em IEEE/CVF Conference on Computer Vision and Pattern Recognition}, pages 1191--1200, 2022.

\bibitem{yin2022content}
Guanghao Yin, Wei Wang, Zehuan Yuan, Chuchu Han, Wei Ji, Shouqian Sun, and Changhu Wang.
\newblock Content-variant reference image quality assessment via knowledge distillation.
\newblock In {\em AAAI}, volume~36, pages 3134--3142, 2022.

\bibitem{zhang2018perceptual}
Richard Zhang, Phillip Isola, Alexei~A Efros, Eli Shechtman, and Oliver Wang.
\newblock The unreasonable effectiveness of deep features as a perceptual metric.
\newblock In {\em IEEE Conference on Computer Vision and Pattern Recognition}, pages 586--595, 2018.

\bibitem{zhang2025evaluating}
Yuhang Zhang, Joshua Maraval, Zhengyu Zhang, Nicolas Ramin, Shishun Tian, and Lu Zhang.
\newblock Evaluating human perception of novel view synthesis: Subjective quality assessment of gaussian splatting and nerf in dynamic scenes.
\newblock {\em arXiv preprint arXiv:2501.08072}, 2025.

\bibitem{zhou2023nerflix}
Kun Zhou, Wenbo Li, Yi Wang, Tao Hu, Nianjuan Jiang, Xiaoguang Han, and Jiangbo Lu.
\newblock Nerflix: High-quality neural view synthesis by learning a degradation-driven inter-viewpoint mixer.
\newblock In {\em IEEE/CVF Conference on Computer Vision and Pattern Recognition}, pages 12363--12374, 2023.

\bibitem{zhu2023DeepDC}
Hanwei Zhu, Baoliang Chen, Lingyu Zhu, Shiqi Wang, and Weisi Lin.
\newblock Deepdc: Deep distance correlation as a perceptual image quality evaluator.
\newblock {\em CoRR}, abs/2211.04927v2, 2023.

\end{thebibliography}
}
\clearpage

\begin{figure*}[h]!
    \centering
    \includegraphics[width=1\textwidth]{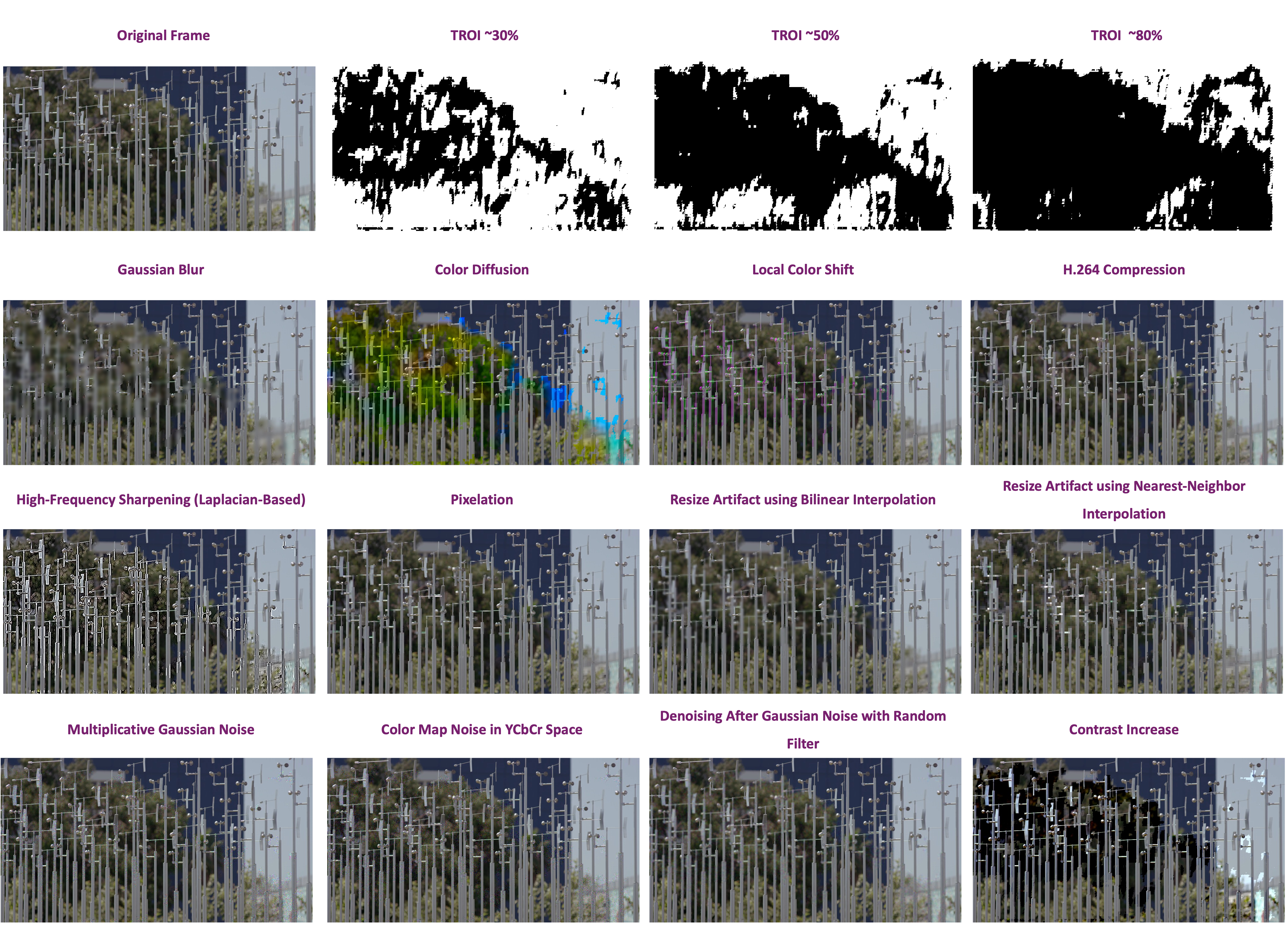}
    \caption{Examples of distortion classes applied to the training dataset. Each row illustrates a different class of synthetic distortion (e.g., blur, color, compression, noise) applied selectively within Temporal Regions of Interest (TROIs) of varying coverage (30\%, 50\%, 80\%) to simulate localized artifacts commonly observed in NVS.}
    \vspace{-0.1in}
    \label{fig:Noises}
\end{figure*}

\section{Supplementary Material}

\subsection{Embedding-Based Visualization of Predicted Distortions}
In this subsection, we provide further explanation of the training dataset with examples of synthetic noise that is added. The training dataset incorporates a broad spectrum of distortion classes designed to reflect realistic and challenging image degradations. These include different types of color, compression, noise, denoising, brightness and illumination, downsampling, sharpening, contrast, and geometric artefacts. Each distortion is applied at five levels of severity to enable fine-grained assessment of model robustness across diverse visual conditions. The distortion were only applied to TROI region only (the black area of TROI in Figure 5). Figure \ref{fig:Noises} illustrates three TROI maps for orgninal image and nine different synthetic artifacts that are applied on TROI 30\% for a frame of video of ``Netflix\_WindAndNature''\footnote{https://media.xiph.org/video/derf/}. 
\subsection{Dense Patch Matching via Nearest-Neighbor Cosine Similarity}

To better understand how our model captures local quality degradations between a reference view and a processed (novel) view, we compute dense patch correspondences using nearest-neighbor cosine similarity. This produces a per-patch \emph{mismatch heatmap} that highlights regions where the processed view deviates most strongly from the reference.

\subsection*{Method}

\begin{enumerate}
    \item \textbf{Patch embeddings.} Both images are divided into $g \times g$ non-overlapping patches and encoded using the NOVA backbone, yielding normalized patch embeddings $\hat{\mathbf{a}}_i, \hat{\mathbf{b}}_j \in \mathbb{R}^C$.
    
    \item \textbf{Similarity matrix.} For every patch pair we compute
    \[
    S_{ij} = \hat{\mathbf{a}}_i^\top \hat{\mathbf{b}}_j .
    \]
    
    \item \textbf{Best matches.} For each patch in the processed image $B$,
    \[
    s^{B\to A}_j = \max_i S_{ij},
    \]
    which records the best-matching similarity to a reference patch.
    
    \item \textbf{Mismatch score.} After min--max normalization, mismatch is defined as the inverted similarity:
    \[
    m^B_j = 1 - \tilde{s}^{B\to A}_j.
    \]
    If the match is not \emph{reciprocal} (i.e., the best match is not mutual), a multiplicative penalty factor $\beta > 1$ is applied.
    
    \item \textbf{Global normalization.} To ensure comparability across both directions, mismatch values from reference and processed images are pooled and globally min--max normalized:
    \[
    m^{B,\text{norm}}_j = \frac{m^B_j - m_{\min}}{m_{\max} - m_{\min} + \varepsilon}.
    \]
    
    \item \textbf{Heatmap construction.} The normalized values $\{m^{B,\text{norm}}_j\}$ are reshaped to the $g \times g$ patch grid and visualized as a heatmap. High intensity values correspond to poorly matched or inconsistent regions in the processed view.
\end{enumerate}

This mismatch heatmap provides a dense, spatially localized measure of how well the processed view preserves the structure and content of the reference. Distortions such as blur, noise, ghosting, or rendering artifacts manifest as elevated mismatch responses. Importantly, because the measure relies on feature correspondences rather than pixel alignment, it is well-suited for NVS scenarios where geometric misalignment is inevitable.

In practice, these heatmaps allow us to visualize and quantify quality differences at a fine-grained level, complementing global quality scores. We include examples for both synthetic distortions (where ground-truth changes are known) and novel view synthesis outputs (where artifacts are scene- and viewpoint-dependent).

\begin{figure*}[t]
    \centering
    \includegraphics[width=1\linewidth]{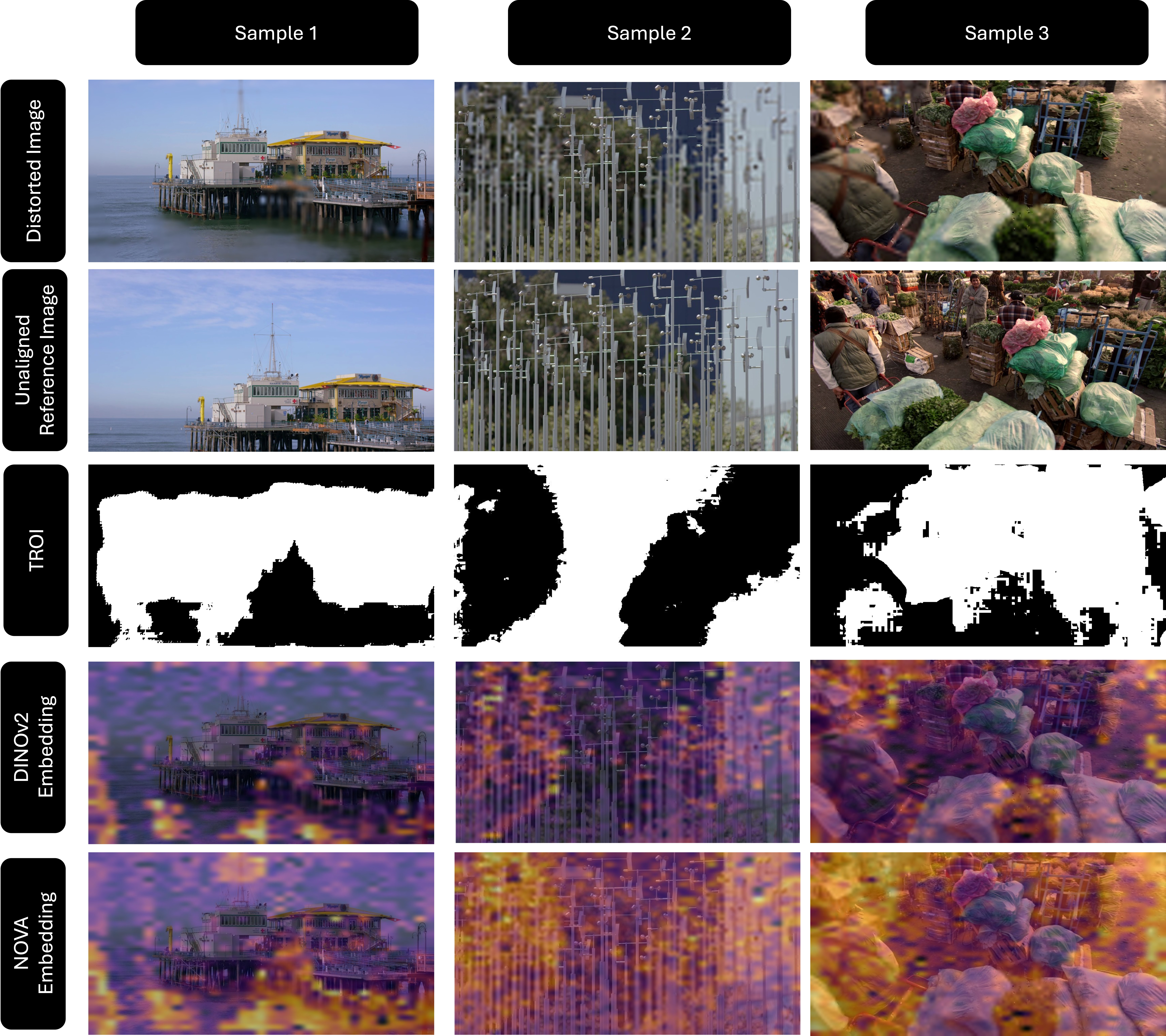}
    \caption{
    Visualization of embedding responses for distorted images relative to unaligned reference images. 
    Each column shows a sample with: (1) the distorted image, (2) the corresponding unaligned reference image, 
    (3) the Temporal Region of Interest (TROI) mask indicating motion-sensitive areas where distortions are most perceptually relevant, 
    (4) heatmaps derived from pretrained DINOv2 embeddings, and 
    (5) heatmaps derived from our proposed NOVA embeddings. 
    Compared to DINOv2, NOVA produces sharper and more localized mismatch responses, especially within TROI regions, 
    highlighting distortions such as structural artifacts and content degradation more faithfully.
    }
    \vspace{-0.1in}
    \label{fig:troi_visualization}
\end{figure*}

\begin{figure*}[t]
    \centering
    \includegraphics[width=1\linewidth]{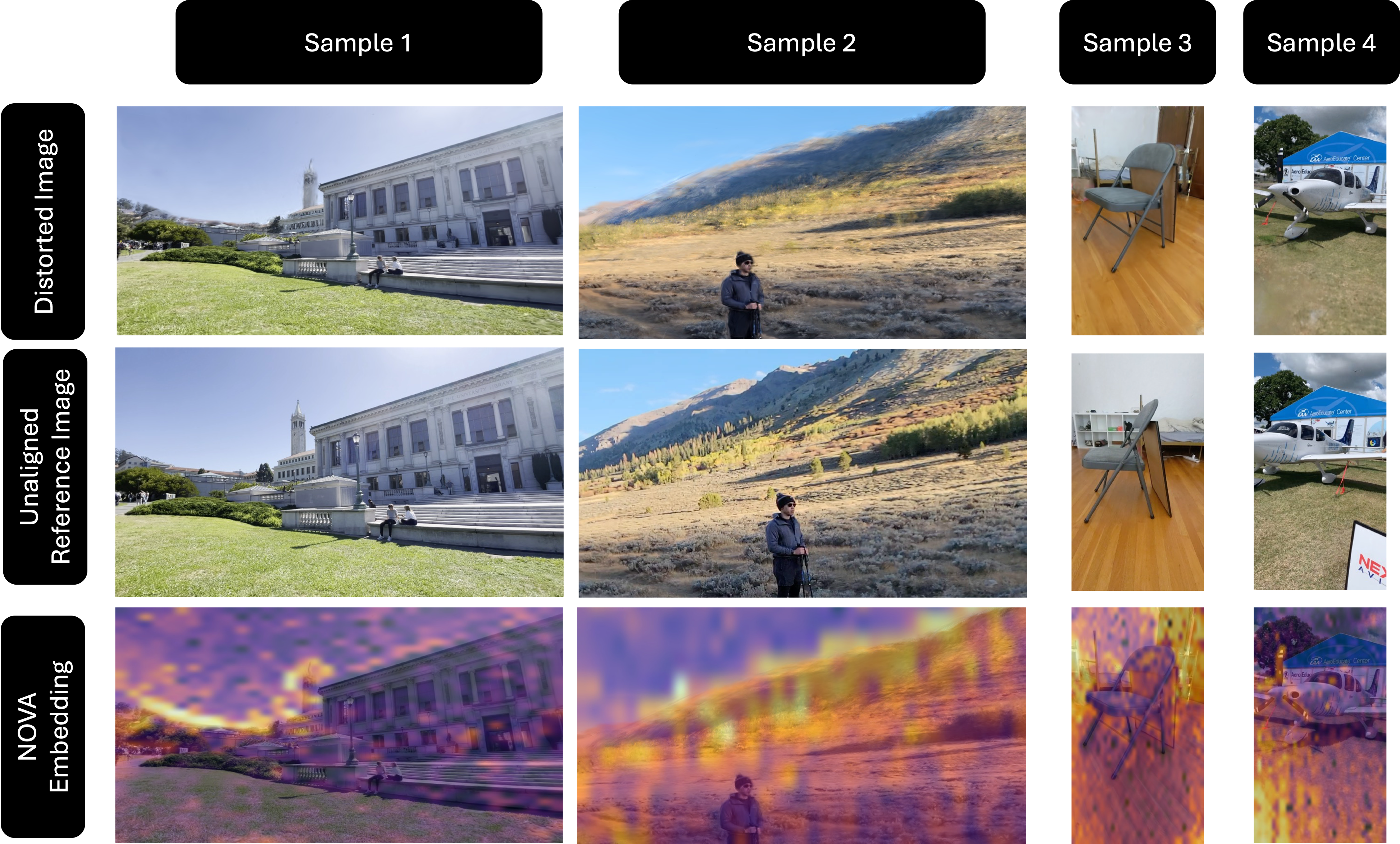}
    \caption{
    Examples from our proposed NVS NAR-IQA benchmark. 
    Each column shows a distorted novel view (top), an unaligned reference image (middle), 
    and the mismatch heatmap derived from our NOVA embeddings (bottom). 
    The heatmaps reveal how distortions in synthesized views; including geometry errors, ghosting, and structural inconsistencies; are highlighted relative to nearby unaligned references. 
    These examples demonstrate that NOVA can capture perceptually relevant degradations in realistic NVS outputs.
    }
    \vspace{-0.1in}
    \label{fig:nvs_examples}
\end{figure*}

To illustrate the effectiveness of our approach, Figures~\ref{fig:troi_visualization} and~\ref{fig:nvs_examples} show qualitative examples for both synthetic distortions and novel view synthesis (NVS) distortions. In the synthetic setting, we include distorted images, unaligned references, TROI masks, and heatmaps from both the pretrained DINOv2 and our NOVA model. While DINOv2 produces diffuse and noisy responses, NOVA yields sharper, more localized mismatch patterns that align closely with the TROI regions, effectively capturing artifacts such as blur, noise, and compression.

In the NVS setting, distorted views are compared with nearby unaligned references, and NOVA highlights structural inconsistencies, ghosting, and texture artifacts characteristic of NVS pipelines. These responses concentrate in perceptually degraded areas, aligning well with human intuition.

Together, the results demonstrate that NOVA not only improves over DINOv2 in localizing synthetic distortions, but also generalizes to realistic NVS artifacts, providing dense, perceptually meaningful mismatch maps for non-aligned reference quality assessment.
% Figure X shows some examples of synthetic distortions and the dense patch matching heatmaps and troi area that the distortion has applied. Our model shows a very good match with troi for different types of distortions. We also show the original Dinov2 

\section{NVS NAR-IQA Benchmark}

To complement Figure~\ref{fig:bench} in the main paper, we present additional scene-wise scores for each IQA model in Table~\ref{tab:bench}. Additionally, we evaluated CrossScore~\cite{wang2024crossscore} using five randomly selected reference frames before and after the aligned ground-truth frame, repeating inference 10 times. The mean score was 59.22 (std. 0.51), whereas our method NOVA achieved 80.19 with a single non-aligned reference in the benchmark. As CrossScore is trained to emulate SSIM, it does not surpass SSIM; however, incorporating multiple nearby references yields measurable improvements and indicates a promising direction for future extensions of NAR-NVSQA. Finally, we emphasize that our model was trained on a large collection of synthetic distortions derived from a video dataset, which does not overlap with our NVS NAR-IQA test dataset.

\begin{table*}[t]
    \resizebox{\textwidth}{!}{
            \centering
            \begin{threeparttable}
            \begin{tabular}
            {@{}lllrrrrrrrrrrrrrrrrrr}
            \toprule
            \rotatebox{90}{Metrics} & \rotatebox{90}{Aligned} & \rotatebox{90}{Non-Aligned} & \rotatebox{90}{Egypt} & \rotatebox{90}{Giannini-Hall} & \rotatebox{90}{aspen} & \rotatebox{90}{campanile} & \rotatebox{90}{desolation} & \rotatebox{90}{dozer} & \rotatebox{90}{floating-tree} & \rotatebox{90}{kitchen} & \rotatebox{90}{library} & \rotatebox{90}{person} & \rotatebox{90}{plane} & \rotatebox{90}{poster} & \rotatebox{90}{redwoods2} & \rotatebox{90}{sculpture} & \rotatebox{90}{storefront} & \rotatebox{90}{stump} & \rotatebox{90}{vegetation} & \rotatebox{90}{Total} \\
            \midrule
            \multicolumn{9}{l}{Full Reference IQA} \\
            \multirow{2}{*}{PSNR} &\checkmark &  & 81.0 & 63.5 & 50.0 & 77.0 & 59.3 & 50.8 & 67.2 & 65.6 & 47.8 & 72.7 & 57.4 & 50.7 & 45.2 & 84.6 & 22.0 & 81.7 & 33.3 & 60.5 \\
             &  &\checkmark & 49.2 & 49.2 & 18.8 & 54.1 & 37.3 & 20.3 & 35.9 & 50.0 & 40.3 & 39.4 & 45.6 & 52.2 & 24.2 & 52.3 & 33.9 & 53.3 & 33.3 & 41.0 \\
             \noalign{\vskip 3pt} % space above the dashed line
            \hdashline
            \noalign{\vskip 3pt} % space below the dashed line
            \multirow{2}{*}{SSIM~\cite{wang2004image}} &\checkmark &  & 81.0 & 65.1 & 57.8 & 77.0 & 61.0 & 55.9 & 65.6 & 68.8 & 37.3 & 69.7 & 61.8 & 38.8 & 41.9 & 90.8 & 32.2 & 75.0 & 37.5 & 60.7 \\
            &  &\checkmark & 69.8 & 47.6 & 31.2 & 45.9 & 47.5 & 30.5 & 39.1 & 57.8 & 50.7 & 50.0 & 44.1 & 61.2 & 29.0 & 80.0 & 44.1 & 36.7 & 37.5 & 47.8 \\
            \noalign{\vskip 3pt} % space above the dashed line
            \hdashline
            \noalign{\vskip 3pt} % space below the dashed line
            LPIPS~\cite{zhang2018perceptual} &\checkmark &  & 71.4 & 71.4 & 65.6 & 80.3 & 74.6 & 55.9 & 78.1 & 85.9 & 67.2 & 72.7 & 79.4 & 50.7 & 45.2 & 86.2 & 40.7 & 83.3 & 37.5 & 68.7 \\
            (Alex) &  &\checkmark & 38.1 & 65.1 & 64.1 & 57.4 & 62.7 & 54.2 & 53.1 & 50.0 & 52.2 & 45.5 & 66.2 & 55.2 & 64.5 & 41.5 & 49.2 & 53.3 & 83.3 & 55.2 \\
            \noalign{\vskip 3pt} % space above the dashed line
            \hdashline
            \noalign{\vskip 3pt} % space below the dashed line
            ST-LPIPS~\cite{ghildyal2022stlpips} &\checkmark &  & 74.6 & 74.6 & 82.8 & 80.3 & 62.7 & 59.3 & 89.1 & 76.6 & 67.2 & 57.6 & 75.0 & 56.7 & 48.4 & 86.2 & 44.1 & 80.0 & 83.3 & 70.1 \\
            (Alex) &  &\checkmark & 49.2 & 50.8 & 62.5 & 52.5 & 54.2 & 55.9 & 45.3 & 31.2 & 49.3 & 47.0 & 58.8 & 47.8 & 54.8 & 40.0 & 47.5 & 41.7 & 83.3 & 50.0 \\
            \noalign{\vskip 3pt} % space above the dashed line
            \hdashline
            \noalign{\vskip 3pt} % space below the dashed line
            LPIPS~\cite{zhang2018perceptual} &\checkmark &  & 88.9 & 73.0 & 62.5 & 78.7 & 67.8 & 59.3 & 81.2 & 90.6 & 67.2 & 75.8 & 72.1 & 46.3 & 40.3 & 87.7 & 37.3 & 88.3 & 37.5 & 69.2 \\
            (VGG) &  &\checkmark & 58.7 & 63.5 & 64.1 & 68.9 & 67.8 & 57.6 & 57.8 & 73.4 & 73.1 & 60.6 & 54.4 & 70.1 & 56.5 & 43.1 & 40.7 & 65.0 & 58.3 & 61.0 \\
            \noalign{\vskip 3pt} % space above the dashed line
            \hdashline
            \noalign{\vskip 3pt} % space below the dashed line
            ST-LPIPS~\cite{ghildyal2022stlpips} &\checkmark &  & 66.7 & 73.0 & 82.8 & 72.1 & 71.2 & 64.4 & 82.8 & 78.1 & 55.2 & 65.2 & 76.5 & 56.7 & 58.1 & 81.5 & 52.5 & 81.7 & 75.0 & 70.0 \\
            (VGG) &  &\checkmark & 47.6 & 49.2 & 59.4 & 44.3 & 52.5 & 66.1 & 54.7 & 45.3 & 41.8 & 45.5 & 63.2 & 56.7 & 64.5 & 43.1 & 54.2 & 45.0 & 83.3 & 52.8 \\
            \noalign{\vskip 3pt} % space above the dashed line
            \hdashline
            \noalign{\vskip 3pt} % space below the dashed line
            LPIPS~\cite{zhang2018perceptual} &\checkmark &  & 82.5 & 73.0 & 75.0 & 80.3 & 74.6 & 62.7 & 81.2 & 85.9 & 67.2 & 81.8 & 79.4 & 53.7 & 41.9 & 86.2 & 44.1 & 86.7 & 41.7 & 71.7 \\
            (Squeeze) &  &\checkmark & 46.0 & 66.7 & 57.8 & 59.0 & 59.3 & 50.8 & 46.9 & 65.6 & 61.2 & 45.5 & 64.7 & 70.1 & 64.5 & 46.2 & 52.5 & 65.0 & 75.0 & 58.1 \\
            \noalign{\vskip 3pt} % space above the dashed line
            \hdashline
            \noalign{\vskip 3pt} % space below the dashed line
            \multirow{1}{*}{DISTS~\cite{Ding20}} &\checkmark &  & 71.4 & 71.4 & 75.0 & 75.4 & 76.3 & 81.4 & 75.0 & 84.4 & 76.1 & 75.8 & 73.5 & 80.6 & 62.9 & 87.7 & 57.6 & 88.3 & 45.8 & 75.2 \\
            &  &\checkmark & 60.3 & 71.4 & 76.6 & 65.6 & 55.9 & 74.6 & 70.3 & 67.2 & 77.6 & 75.8 & 73.5 & 83.6 & 66.1 & 76.9 & 67.8 & 73.3 & 66.7 & 71.1 \\
            \noalign{\vskip 3pt} % space above the dashed line
            \hdashline
            \noalign{\vskip 3pt} % space below the dashed line
            \multirow{2}{*}{FLIP~\cite{andersson2020flip}}&\checkmark &  & 84.1 & 68.3 & 50.0 & 57.4 & 57.6 & 47.5 & 65.6 & 57.8 & 49.3 & 60.6 & 57.4 & 55.2 & 53.2 & 84.6 & 32.2 & 80.0 & 33.3 & 59.5 \\
            &  &\checkmark & 46.0 & 54.0 & 31.2 & 36.1 & 49.2 & 27.1 & 50.0 & 59.4 & 53.7 & 42.4 & 42.6 & 58.2 & 33.9 & 61.5 & 33.9 & 43.3 & 50.0 & 45.5 \\
            \noalign{\vskip 3pt} % space above the dashed line
            \hdashline
            \noalign{\vskip 3pt} % space below the dashed line
            \multirow{2}{*}{DeepDC~\cite{zhu2023DeepDC}} &\checkmark &  & 71.4 & 79.4 & 78.1 & 78.7 & 71.2 & 79.7 & 84.4 & 82.8 & 67.2 & 80.3 & 77.9 & 82.1 & 71.0 & 87.7 & 64.4 & 86.7 & 75.0 & 77.7 \\
            &  &\checkmark & 54.0 & 71.4 & 73.4 & 68.9 & 72.9 & 83.1 & 79.7 & 70.3 & 61.2 & 80.3 & 67.6 & 77.6 & 75.8 & 75.4 & 72.9 & 71.7 & 66.7 & 72.1 \\
            \noalign{\vskip 3pt} % space above the dashed line
            \hdashline
            \noalign{\vskip 3pt} % space below the dashed line
            % ZS-IQA (L2, CLIP-ViT-B/32) & check &  & 90.5 & 74.6 & 59.4 & 82.0 & 71.2 & 57.6 & 78.1 & 89.1 & 71.6 & 77.3 & 67.6 & 37.3 & 51.6 & 86.2 & 40.7 & 90.0 & 45.8 & 69.8 \\
            % ZS-IQA (L2, CLIP-ViT-B/32) &  & check & 61.9 & 46.0 & 48.4 & 70.5 & 52.5 & 32.2 & 40.6 & 35.9 & 55.2 & 59.1 & 44.1 & 59.7 & 54.8 & 41.5 & 50.8 & 51.7 & 62.5 & 50.6 \\
            ZS-IQA~\cite{ghildyal2025foundation} & \checkmark &  & 88.9 & 71.4 & 62.5 & 80.3 & 74.6 & 59.3 & 79.7 & 93.8 & 73.1 & 77.3 & 64.7 & 40.3 & 50.0 & 87.7 & 37.3 & 86.7 & 33.3 & 69.7 \\
            (L2, Dinov1) &  & \checkmark & 42.9 & 57.1 & 51.6 & 67.2 & 44.1 & 32.2 & 43.8 & 48.4 & 62.7 & 56.1 & 50.0 & 49.3 & 59.7 & 43.1 & 44.1 & 41.7 & 58.3 & 50.0 \\
            \noalign{\vskip 3pt} % space above the dashed line
            \hdashline
            \noalign{\vskip 3pt} % space below the dashed line
            % ZS-IQA (Cos, CLIP-ViT-B/32) & check &  & 87.3 & 74.6 & 62.5 & 82.0 & 69.5 & 61.0 & 82.8 & 89.1 & 70.1 & 75.8 & 67.6 & 34.3 & 41.9 & 87.7 & 33.9 & 86.7 & 45.8 & 68.7 \\
            % ZS-IQA (Cos, CLIP-ViT-B/32) &  & check & 66.7 & 60.3 & 45.3 & 60.7 & 50.8 & 42.4 & 46.9 & 59.4 & 61.2 & 62.1 & 52.9 & 56.7 & 50.0 & 49.2 & 42.4 & 40.0 & 66.7 & 53.4 \\
            ZS-IQA~\cite{ghildyal2025foundation} & \checkmark &  & 88.9 & 71.4 & 60.9 & 80.3 & 69.5 & 61.0 & 81.2 & 93.8 & 70.1 & 74.2 & 63.2 & 41.8 & 43.5 & 89.2 & 39.0 & 90.0 & 29.2 & 69.0 \\
            (Cos, Dinov1) &  & \checkmark & 55.6 & 63.5 & 56.2 & 67.2 & 62.7 & 45.8 & 57.8 & 65.6 & 67.2 & 63.6 & 58.8 & 55.2 & 56.5 & 40.0 & 47.5 & 56.7 & 66.7 & 57.8 \\

            \noalign{\vskip 3pt} % space above the dashed line
            \hdashline
            \noalign{\vskip 3pt} 
            \multirow{2}{*}{DreamSim \cite{fu2024dreamsim}} &\checkmark &  & 63.5 & 68.3 & 43.8 & 68.9 & 55.9 & 45.8 & 64.1 & 65.6 & 53.7 & 48.5 & 57.4 & 34.3 & 37.1 & 76.9 & 59.3 & 70.0 & 54.2 & 56.9 \\
             &  &\checkmark & 20.6 & 20.6 & 35.9 & 19.7 & 30.5 & 42.4 & 28.1 & 10.9 & 20.9 & 10.6 & 26.5 & 52.2 & 43.6 & 13.9 & 44.1 & 15.0 & 37.5 & 27.3 \\

            \midrule
            \multicolumn{9}{l}{No Reference IQA} \\
            ARNIQA~\cite{agnolucci2024arniqa} &  &  & 38.1 & 44.4 & 57.8 & 41.0 & 55.9 & 42.4 & 60.9 & 67.2 & 49.2 & 51.5 & 55.9 & 64.2 & 71.0 & 46.2 & 69.5 & 58.3 & 62.5 & 55.1 \\
            CONTRIQUE~\cite{madhusudana2022image} &  &  & 42.9 & 69.8 & 60.9 & 37.7 & 52.5 & 52.5 & 60.9 & 59.4 & 47.8 & 50.0 & 61.8 & 52.2 & 64.5 & 66.1 & 47.5 & 80.0 & 41.7 & 55.8 \\
            LAR-IQA~\cite{jamshidi2025lar} &  &  & 42.9 & 39.7 & 43.8 & 22.9 & 49.1 & 44.1 & 62.5 & 50.0 & 49.2 & 39.4 & 55.9 & 77.6 & 77.4 & 35.4 & 84.8 & 51.7 & 54.2 & 51.8 \\
            MANIQA~\cite{yang2022maniqa} &  &  & 34.9 & 55.6 & 64.1 & 32.8 & 64.4 & 54.2 & 60.9 & 65.6 & 43.3 & 57.6 & 69.1 & 61.2 & 61.3 & 41.5 & 62.7 & 60.0 & 83.3 & 57.2 \\
            HyperIQA~\cite{Su_2020_CVPR} &  &  & 38.1 & 42.9 & 71.9 & 32.8 & 61.0 & 61.0 & 54.7 & 64.1 & 53.7 & 56.1 & 63.2 & 68.7 & 70.9 & 40.0 & 67.8 & 61.7 & 66.7 & 57.0 \\
            GraphIQA~\cite{sun2022graphiqa} & & & 55.6 & 71.4 & 78.1 & 36.1 & 72.9 & 71.2 & 68.8 & 78.1 & 56.7 & 72.7 & 75.0 & 73.1 & 80.6 & 78.5 & 76.3 & 78.3 & 91.7 & 54.5 \\
            TRIQA~\cite{sureddi2025triqa} & & & 58.3 & 58.7 & 60.5 & 58.1 & 57.1 & 61.0 & 59.4 & 59.4 & 59.7 & 58.6 & 58.6 & 58.6 & 58.6 & 58.6 & 58.6 & 58.6 & 58.6 & 58.3 \\
            AGAIQA~\cite{wang2024blind} & & & 55.6 & 71.4 & 78.1 & 36.1 & 72.9 & 71.2 & 68.8 & 78.1 & 56.7 & 72.7 & 75.0 & 73.1 & 80.6 & 78.5 & 76.3 & 78.3 & 91.7 & 70.7 \\
            NVS-SQA \cite{qu2025nvs} &  &  & 33.3 & 66.7 & 75.0 & 21.3 & 69.5 & 62.7 & 67.2 & 48.4 & 53.7 & 48.5 & 64.7 & 53.7 & 74.2 & 29.2 & 54.2 & 73.3 & 75.0 & 56.3 \\

            \midrule
            \multicolumn{9}{l}{Non-Aligned Reference IQA} \\
            \multirow{2}{*}{CVRKD~\cite{yin2022content}} &\checkmark &  & 60.3 & 60.3 & 50.0 & 70.5 & 47.5 & 50.8 & 43.8 & 32.8 & 52.2 & 59.1 & 44.1 & 29.9 & 46.8 & 47.7 & 42.4 & 50.0 & 50.0 & 49.2 \\
            &  &\checkmark & 65.1 & 58.7 & 54.7 & 60.7 & 55.9 & 44.1 & 51.6 & 43.8 & 53.7 & 60.6 & 36.8 & 46.3 & 46.8 & 44.6 & 55.9 & 48.3 & 37.5 & 51.3 \\
            \noalign{\vskip 3pt} % space above the dashed line
            \hdashline
            \noalign{\vskip 3pt} % space below the dashed line
            \multirow{2}{*}{CrossScore~\cite{wang2024crossscore}} &\checkmark &  & 76.2 & 44.4 & 51.6 & 29.5 & 55.9 & 61.0 & 67.2 & 70.3 & 52.2 & 34.8 & 67.6 & 76.1 & 72.6 & 33.8 & 59.3 & 48.3 & 62.5 & 56.5 \\
            &  &\checkmark & 71.4 & 41.3 & 48.4 & 26.2 & 59.3 & 55.9 & 67.2 & 60.9 & 62.7 & 51.5 & 70.6 & 65.7 & 62.9 & 38.5 & 62.7 & 45.0 & 79.2 & 56.3 \\
            \noalign{\vskip 3pt} % space above the dashed line
            \hdashline
            \noalign{\vskip 3pt} % space below the dashed line
            \textbf{NOVA} &\checkmark &  & 84.1 & 85.7 & 79.7 & 65.6 & 76.3 & 79.7 & 90.6 & 89.1 & 83.6 & 78.8 & 82.3 & 94.0 & 71.0 & 93.8 & 64.4 & 85.0 & 70.8 & \textbf{81.45} \\
            \textbf{(ours)} &  &\checkmark & 82.5 & 84.1 & 85.9 & 65.6 & 69.5 & 76.3 & 90.6 & 90.6 & 83.6 & 80.3 & 77.9 & 95.5 & 67.7 & 87.7 & 66.1 & 78.3 & 70.8 & \textbf{80.19} \\

            \bottomrule
        
        \end{tabular}
        \end{threeparttable}
    }
    \vspace{-0.1in}
    \caption{
    Quantitative comparison (in \% accuracy) of IQA models on the NVS NAR-IQA benchmark across 17 diverse scenes. We evaluate full-reference, non-aligned reference, and no-reference metrics under both aligned and non-aligned reference conditions.}
    \label{tab:bench}
    \vspace{-0.2in}
\end{table*}

\end{document}